\pdfoutput=1

\documentclass[11pt]{article}

\usepackage[final]{acl}

\usepackage{times}
\usepackage{latexsym}

\usepackage[T1]{fontenc}

\usepackage[utf8]{inputenc}

\usepackage{microtype}

\usepackage{inconsolata}

\usepackage{xcolor}
\usepackage{amsmath}

\usepackage{hhline}
\usepackage{url}
\usepackage{hhline}
\usepackage{url}
\usepackage{comment}
\usepackage{threeparttable}
\usepackage{textcomp}
\usepackage{parskip}
\usepackage{hhline}
\usepackage{adjustbox}
\usepackage{fixfoot}
\usepackage{enumerate}
\usepackage{lscape} 
\usepackage{booktabs}
\usepackage{multirow}
\usepackage{color,soul}
\sethlcolor{yellow}
\usepackage{rotating}

\title{Adverse Event Extraction from Discharge Summaries: A New Dataset, Annotation Scheme, and Initial Findings}

\author{
 \textbf{Imane Guellil\textsuperscript{1}},
 \textbf{Salomé Andres\textsuperscript{1}},
 \textbf{Atul Anand\textsuperscript{1}},  \textbf{Bruce Guthrie\textsuperscript{1}},
\\
  \textbf{Huayu Zhang\textsuperscript{1}},
  \textbf{Abul Hasan\textsuperscript{2}},
  \textbf{Honghan Wu\textsuperscript{3,2}},
  \textbf{Beatrice Alex\textsuperscript{1}},
\\
  \textsuperscript{1}University of Edinburgh,
  \textsuperscript{2}University College London,
  \textsuperscript{3}University of Glasgow,
\\
  \small{
    \textbf{Correspondence:} \href{mailto:imane.guellil.ig@gmail.com}{imane.guellil.ig@gmail.com}
  }
}

\begin{document}
\maketitle
\begin{abstract}
In this work, we present a manually annotated corpus for Adverse Event (AE) extraction from discharge summaries of elderly patients, a population often underrepresented in clinical NLP resources. The dataset includes 14 clinically significant AEs—such as falls, delirium, and intracranial haemorrhage, along with contextual attributes like negation, diagnosis type, and in-hospital occurrence. Uniquely, the annotation schema supports both discontinuous and overlapping entities, addressing challenges rarely tackled in prior work. We evaluate multiple models using FlairNLP across three annotation granularities: fine-grained, coarse-grained, and coarse-grained with negation. While transformer-based models (e.g., BERT-cased) achieve strong performance on document-level coarse-grained extraction (F1 = 0.943), performance drops notably for fine-grained entity-level tasks (e.g., F1 = 0.675), particularly for rare events and complex attributes. These results demonstrate that despite high-level scores, significant challenges remain in detecting underrepresented AEs and capturing nuanced clinical language. Developed within a Trusted Research Environment (TRE), the dataset is available upon request via DataLoch and serves as a robust benchmark for evaluating AE extraction methods and supporting future cross-dataset generalisation.
\end{abstract}
\setcounter{figure}{0}
\renewcommand{\figurename}{Table}
\section{Introduction}
The automatic identification and extraction of Adverse Events (AEs) from clinical texts is a critical area of research in Natural Language Processing (NLP), with important applications in pharmacovigilance and patient safety. AEs refer to adverse outcomes caused by medical interventions, and their accurate detection is essential for supporting clinical decision-making and improving healthcare monitoring systems. 
\setcounter{figure}{0}
\renewcommand{\figurename}{Figure}
\begin{figure}[h]
    \includegraphics[width=1\linewidth]{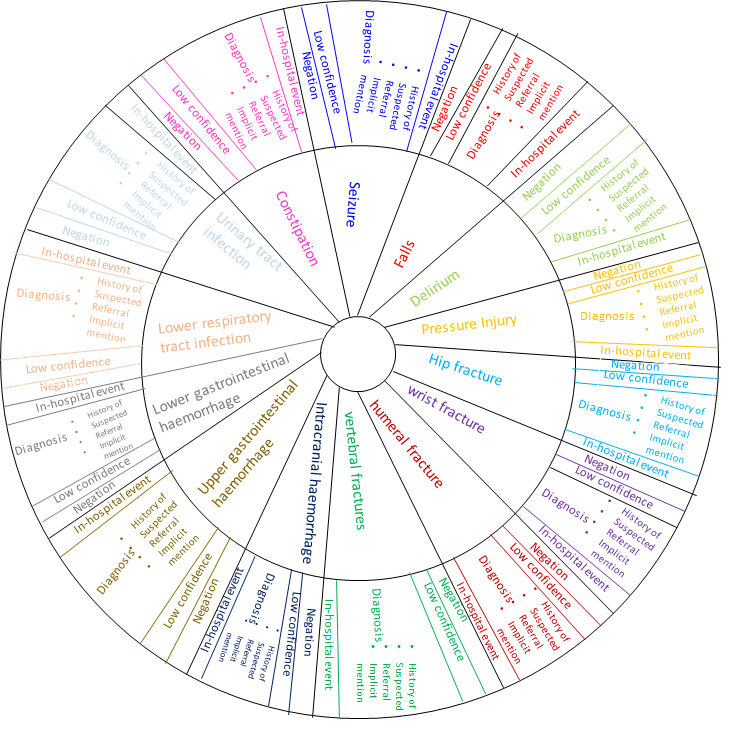}
    
    \caption{AE entities and attributes}
    \label{granularity}
\end{figure}

\newpage
Among clinical documents, discharge summaries-particularly those concerning elderly patients-pose unique challenges due to the complexity, variability, and ambiguity of their narrative structure \cite{murphy2023adverse}.

While recent advances in machine learning, particularly transformer-based models, have improved AE extraction, model performance remains heavily dependent on the availability of high-quality annotated corpora \cite{mahendran2021extracting}. In practice, however, such datasets are limited, especially those tailored to clinical narratives, and existing resources often fail to address key linguistic challenges such as discontinuous and overlapping named entities. These phenomena are especially prevalent in clinical texts, where the structure of medical language can result in multi-part, non-contiguous mentions or nested concepts. For instance, the sentence "The wrist X-ray performed in A\&E confirms the presence of a fracture" includes a discontinuous mention ("wrist" and "fracture") referring to a single AE. Similarly, overlapping entities arise when a shared span belongs to more than one annotation, as in "urinary and faecal incontinence", where "incontinence" is shared between two distinct AE labels.
Such complex cases challenge standard Named Entity Recognition (NER) models, which were originally developed for flat entity structures and are often ill-equipped to handle biomedical texts that demand more expressive annotation. As the field evolves toward unified NER—encompassing flat, overlapping, and discontinuous structures—clinical NLP must adopt more sophisticated annotation schemas and evaluation strategies to reflect real-world use cases.
To address these limitations, we present a manually annotated corpus of discharge summaries from elderly patients, in which AEs are systematically identified and labeled using a clinically informed annotation guideline. Our corpus includes 14 clinically significant AEs (e.g., falls, delirium, hip fractures, urinary tract infections, intracranial haemorrhages), each annotated with four contextual attributes: negation, in-hospital occurrence, diagnosis type (history\_of, suspected, referral, implicit\_mention), and low confidence. The annotation supports both fine-grained and coarse-grained extraction, and explicitly incorporates discontinuous and overlapping entity spans. The complete entity-attribute schema is shown in Figure~\ref{granularity}.

To benchmark this dataset, we evaluate multiple pre-trained language models using the FlairNLP framework, including BioBERT and variants of BERT, across multiple annotation granularities. Our goal is to create a practical and challenging evaluation framework that mirrors the real-world conditions and annotation challenges encountered in clinical EHR data.. Our dataset is available upon request via DataLoch, and offers a valuable foundation for future work in AE extraction, dataset generalization, and unified NER in clinical NLP.

\section{Related work}
The extraction of adverse events (AEs) using NLP has become a crucial area of research due to its potential to improve pharmacovigilance and enhance patient safety \cite{guellil2024natural}. Much of the research in this domain can be categorised into three primary areas: AE classification \cite{tafti2017adverse,ellenius2017detecting,kim2018machine,chen2019detecting,ujiie2020identification}, AE extraction\cite{munkhdalai2018clinical,wu2021chinese,portelli2021improving,zhang2021identifying,yahya2022automatic,li2024improving} and AE normalisation (or mapping) \cite{emadzadeh2018hybrid,combi2018normalizing}. Some papers address multiple tasks simultaneously, such as classification combined with extraction \cite{shen2021automatic}, extraction paired with relation extraction \cite{florez2019deep}, or a combination of classification, extraction and RE \cite{yang2019madex}. Others focus on the severity of AEs and the relationship between AEs and their seriousness \cite{lavertu2021quantifying,d2023biodex}. 

The training of classification, extraction, and normalisation models relies on the availability of a consistently reliable annotated corpus, which remains a fundamental priority. Most of the datasets used come from publicly available resources, primarily sourced from social media platforms and shared tasks. Social media platforms, particularly Twitter, have emerged as a rich source of AE-related data due to users’ frequent sharing of personal experiences with drug-related side effects \cite{ellenius2017detecting,portelli2021improving,zhang2021identifying,guellil2022edinburgh_ucl_health,emadzadeh2018hybrid,li2024improving}. The Social Media Mining for Health (SMM4H) dataset, which contains tweets annotated for AE classification, extraction, and normalisation, and other shared task datasets (such as MADE and n2c2) has become one of the most widely used resources in AE detection studies \cite{yang2019madex,florez2019deep,chen2019extraction,portelli2021improving}. In addition to social media data, clinical datasets such as SIDER\footnote{SIDER contains information on marketed medicines and their recorded adverse drug reactions}, FAERS\footnote{FAERS: FDA Adverse Event Reporting System}, VAERS\footnote{Vaccine Adverse Event Reorting System} (or KAERS for the Korean system) or those from pharmacovigilance databases (in English or other languages) \cite{kim2018machine,combi2018normalizing,chen2019detecting,ujiie2020identification,shen2021automatic,lavertu2021quantifying,li2024improving} have also been widely used in AE-related research. However, we also observe a scarcity of studies that focus on electronic health records \cite{yang2019madex,wu2021chinese} for the extraction of AEs.

The majority of datasets used in previous studies have been manually annotated, often involving two or three annotators to ensure reliability \cite{tafti2017adverse,emadzadeh2018hybrid,ujiie2020identification,shen2021automatic}. Inter-annotator agreement (IAA) is commonly measured using Cohen's kappa \cite{cohen1960coefficient} for classification tasks \cite{tafti2017adverse,chen2019detecting,ujiie2020identification}, while the F1-score is predominantly used for evaluating extraction tasks \cite{munkhdalai2018clinical,d2023biodex}. Only a limited number of studies consider more complex annotation challenges, such as discontinuous and overlapping entities \cite{kim2018machine,yang2019madex}, which can significantly impact the quality and usability of annotated data. Furthermore, the consistency of manual annotation largely depends on the availability and clarity of the annotation guidelines. Despite their crucial role in ensuring coherence, only a few studies explicitly present these guidelines \cite{,henry20202018,chen2019detecting,yang2019madex}, making it difficult to assess the reproducibility and robustness of the annotation process.

\subsection{Contribution}
The use of NLP for detecting AEs has advanced significantly, with notable improvements in classification, Named Entity Recognition (NER), normalization and relation Extraction (RE). Transformer-based models, such as BERT and BioBERT, have played a key role in these advancements, enhancing performance across various AE-related tasks. However, challenges persist, particularly in addressing the varying severity of AEs. In clinical practice, not all AEs carry the same level of urgency - some require immediate medical intervention, while others, though less critical, still have a considerable impact. For instance, conditions like intracranial haemorrhage demand urgent attention, whereas issues such as constipation, while important, are less pressing.

To address these challenges, our work focuses on detecting 14 specific AEs, categorized by their urgency. These range from life-threatening events, such as intracranial haemorrhage and lower respiratory tract infections, to less critical conditions. We manually annotated a corpus of 2,040 discharge summaries (provided by Dataloch\footnote{Dataloch: a Scottish regional trusted research environment (TRE)}) to identify these AEs, ensuring both accuracy and consistency through a carefully developed annotation guideline created in collaboration with clinicians. To promote further research in this area, both the annotation guidelines and the dataset have been made publicly available (upon request from Dataloch). Beyond AE detection, our system also predicts additional attributes, such as whether the event occurred in the hospital, whether it is affirmative or negated and other relevant details. Unlike many existing studies, we pay particular attention to discontinuous and overlapping events, which are often overlooked in previous research. To validate our dataset, we experiment with and compare a variety of model embeddings, ensuring a comprehensive evaluation of their effectiveness. For a better comparison of our contribution to the proposed studies, please refer to table \ref{relatedwork} in Appendix \ref{table_related}.

\section{Annotation guidelines}

There is no consensus on an exhaustive list of AEs. In the context of this study, we reviewed the current literature and organised a Public Partipation Involvement and Engagement (PPIE) workshop for older adults to propose a patient-orientated NLP tool. Based on the information collected, we collaborated with an expert panel of clinicians using a consensus approach to select 14 AEs (detailed in section \ref{ann_labels}). We selected these AEs because they are both common and have a significan impact on patients' quality of life, often requiring substantial medical intervention.

\subsection{Annotation labels}
\label{ann_labels}

In the context of this work, each type of AE represents a label. The following list provides an example for each of the different types:

\begin{enumerate}
    \item \textbf{Fall:} "Three \hl{falls} in last yr"

    \item \textbf{Delirium:} "On admission, she was \hl{confused}. This settled to some extent with pain relief and rehydration."

    \item \textbf{Pressure injury:} "Noted \hl{decubitus ulcer} on the right heel"
    
    \item \textbf{Hip fracture:} "Admitted for a left \hl{intertrochanteric fracture}"

    \item \textbf{Wrist fracture} "\hl{Wrist fracture} shown on the X-ray." 

    \item \textbf{Proximal humeral fracture:} "He sustained a neck of \hl{humerus fracture} on the left side a year ago"

     \item \textbf{Vertebral compression fracture:} "She has several known osteoporotic \hl{vertebral fractures}"  

     \item \textbf{Intracranial haemorrhage:}  "Right parietal \hl{subdural haematoma} visible on CT"

     \item \textbf{Upper gastrointestinal haemorrhage:}  "Patient admitted to A\&E with 3 episodes of \hl{haematemesis}."

     \item \textbf{Lower gastrointestinal Haemorrhage:} "She reported some \hl{PR bleed}" 
     
     \item \textbf{Lower respiratory tract infection:} "New cough and wheeze during admission. Treated as hospital-acquired \hl{pneumonia}"

      \item \textbf{Urinary tract infection:} "CRP was raised, urine dip was positive for blood and leukocytes and Ciprofloxacin was started for \hl{UTI}." 

      \item \textbf{Constipation:} "Patient had \hl{difficulties voiding} despite laxatives. This was resolved with an enema." 

      \item \textbf{Seizure:} "Found by the nurse with generalised \hl{tonic-clonic episode}"

\end{enumerate}

\setcounter{figure}{1}
\renewcommand{\figurename}{Figure}
\begin{figure*}[h]
    \includegraphics[width=1\linewidth]{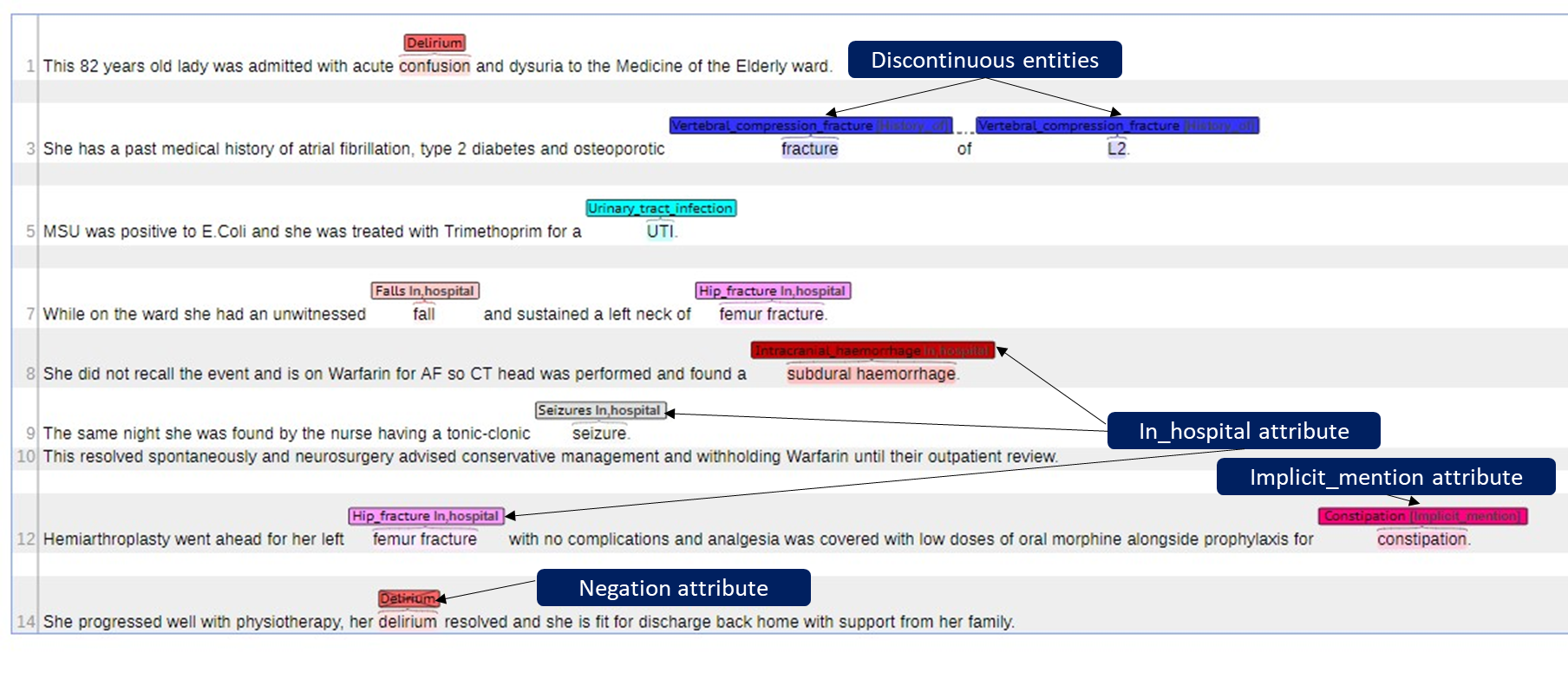}
    
    \caption{A synthetic example annotated in Brat includes discontinuous entities and different attributes.}
    \label{example_all}
\end{figure*}

\subsection{Annotation attributes}
Annotation attributes are additional features or characteristics associated with the tag. We use four main attributes: Negation, low confidence, in-hospital events and Diagnosis. \textit{Negation} can be formulated explicitly with "no" or "not" but can also appear as acronyms (e.g. NAD, meaning nothing abnormal discovered) or be inferred from the text. The annotator did when possible, not include the negation word(s) in the selection of text but simply highlight the words which reference the AE and add the negation attribute to the label (eg. "She slipped but did not \hl{fall}." - This should be annotated \textit{Falls} with the negation attribute.). The \textit{Low confidence} should be selected when the annotator is uncertain of their choice (e.g. "She was admitted with delirium, was confused in the ambulance and had visual \hl{hallucinations} at home. Her daughter said she was describing children running around but they were by themselves in the house. The delirium resolved with rehydration.") - Hallucinations are rarely present in delirium but it seems to be described by the author of the letter as such. The \texttt{in-hospital\_event} attribute allows to specify that the onset of AE took place in the hospital (e.g.~"New cough and wheeze during admission. Treated as hospital-acquired \hl{pneumonia}".

The \textit{Diagnosis} attribute allows the annotator to further qualify the annotated text when the AE is not directly present in the patient's admission, attendance or report. For this purpose, four tags are available:
\begin{itemize}
    \item \textit{History of}, this can be mentioned in the patient's past medical history or something which happened before the admission or attendance (eg. "PMH: T2DM, Parkinson's disease, \hl{subdural haemorrhage}." - where PMH reffers to the past medical history)
    \item \textit{Suspected}, used when the writer of the clinical document suspects the patient has an AE. This attribute can also be used in annotating screening tests which require further investigations to confirm a diagnosis (eg. "This patient was admitted following a GP visit who suspects she might have a \hl{wrist fracture}.")
    \item \textit{Referral} qualifies when a patient is being referred to a different healthcare professional (eg."His haemoglobin remained stable and was referred to the GI team to discuss outpatient investigations of this haematemesis"
    \item \textit{Implicit mention} qualifies an AE which is mentioned in the text but does not apply to the patient. This includes descriptions of the patient's family history or diagnosis given to their spouse or risks of an AE happening (e.g. risk of falls). For example: "He was given laxatives for prophylaxis of \hl{constipation}" - The term prophylaxis means the patient was given laxatives as a preventative measure.
\end{itemize}

\begin{figure}[t]
    \includegraphics[width=1\linewidth]{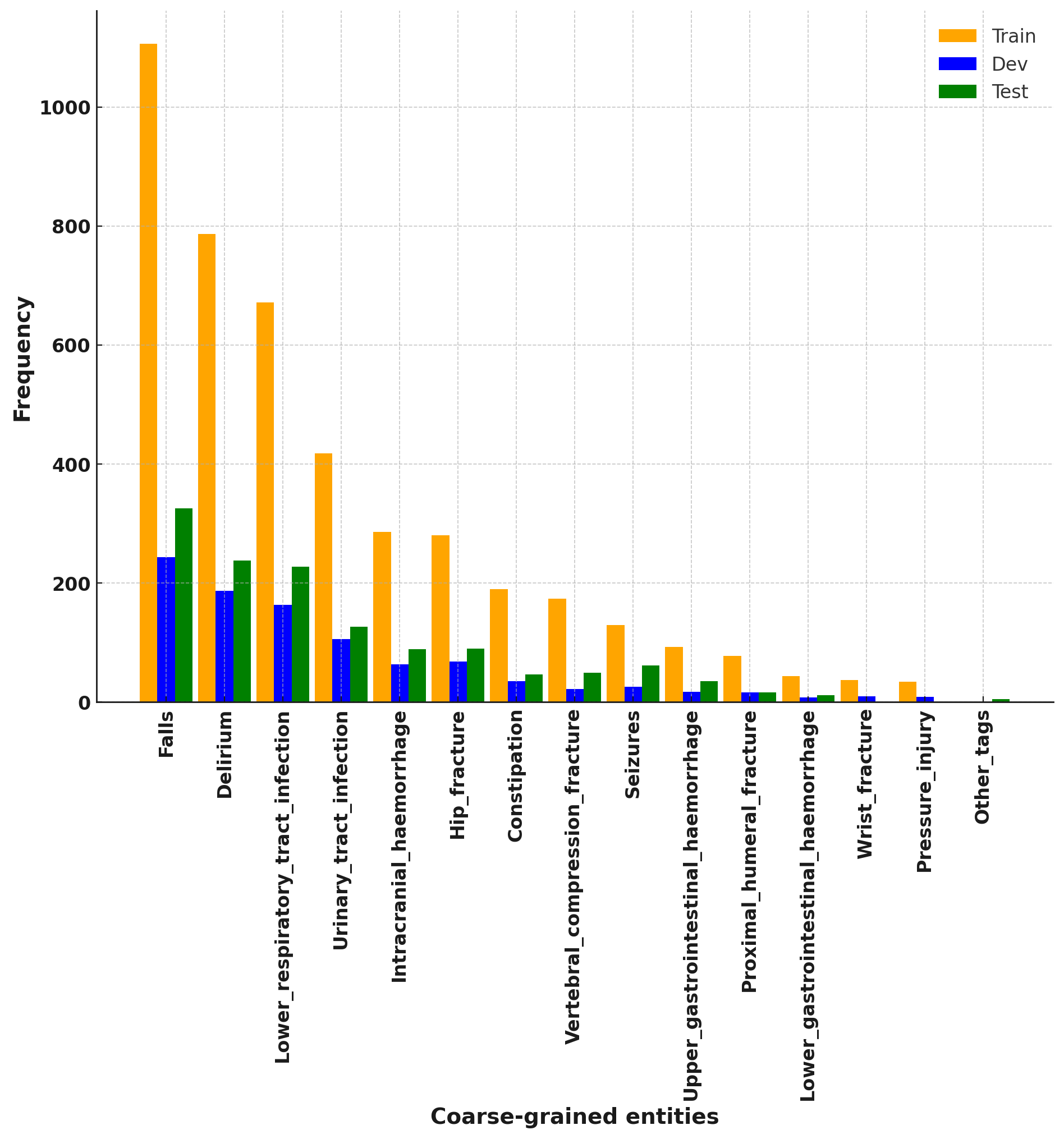}
    
    \caption{The distribution of coarse-grained entities in the train, dev and test datasets}
    \label{coarse_grained_stat}
\end{figure}

\begin{figure}[t]
    \includegraphics[width=1\linewidth]{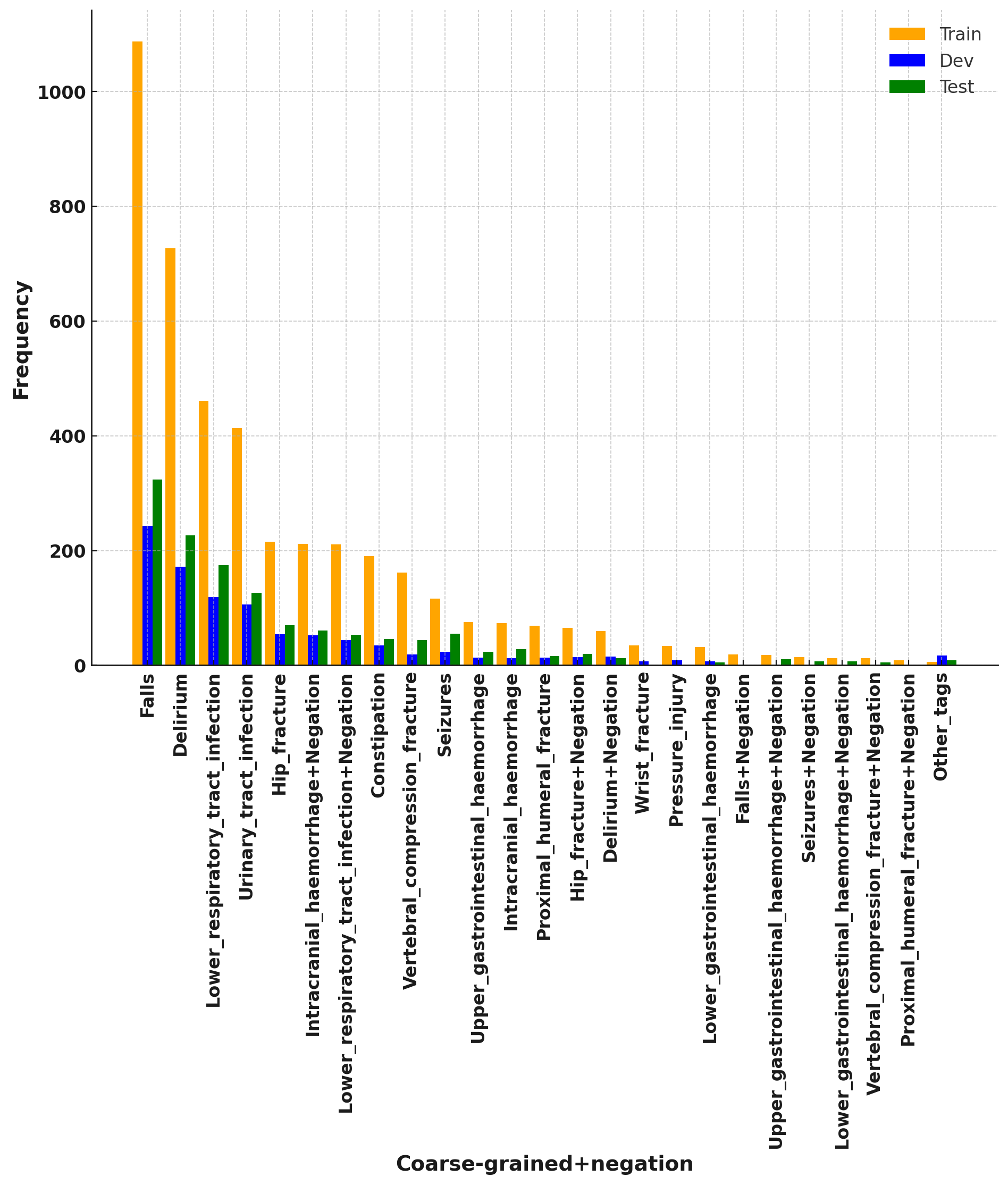}
    
    \caption{The distribution of coarse-grained+negation entities in the train, dev and test datasets}
    \label{coarse_grained_negation_stat}
\end{figure}

\subsection{Discontinuous and overlapping entities:} 
Medical texts often contain distant entities which need to be annotated together to make sense of the label which is being recognised. Discontinuous entities, also named distant entities or fragmented entities, are often used for this purpose \cite{huang2023t}. For example:

\begin{itemize}
    \item “The \hl{wrist} X-ray performed in A\&E confirms the presence of a \hl{fracture}” - These two distant tokens should be annotated as one discontinuous entity with the label \textit{Wrist fracture}
\end{itemize}

This does not apply when annotating two different, isolated symptoms which are associated with the same AE. For example:

\begin{itemize}
    \item “She was \hl{confused} and \hl{agitated} on the ward” - since both of these symptoms, in isolation, are sufficient to detect delirium, the tokens highlighted in this example must be annotated as two separate entities with the label \textit{Delirium}. 
\end{itemize}

Medical texts also tend to contain overlapping entities which require the annotator to highlight the same element of text twice and this can also be used in combination with the distant entities \cite{huang2023t}. For example: 

“When she fell, she \hl{broke} her \hl{wrist} and her \hl{hip}” - This can be annotated with “broke” + “wrist” for the label \textit{Wrist fracture} and “broke” + “hip” for the label \textit{hip fracture}.

Due to space constraint, we present in this paper the most important points related to our guideline. The complete version of the guideline proposed to the annotators is presented in the appendix \ref{detailedguideline}.

\setcounter{figure}{0}
\renewcommand{\figurename}{Table}
\begin{figure}[h]
\begin{center}
     \begin{adjustbox}{width=0.5\textwidth}

\begin{threeparttable}
\caption{IAA between the two annotators (entity and document levels) using bratiaa and Cohen's kappa}
   \label{iaa_entity_doc}

\begin{tabular}{|p{1cm}|p{2.5cm}|p{1cm}|p{1cm}|p{1.5cm}|}
\hline
    \multirow{2}{*}{Corpus}&\multirow{2}{*}{Annotation type}&Entity level&\multicolumn{2}{|c|}{Document level}\\
   \cline{3-5}
   &&F1-score&F1-score&Cohen's kappa\\
   \hline
    \multirow{3}{*}{Pilot}&Fine-grained&0.532&0.720&0.714\\
    \cline{2-5}
    &coarse-grained&0.675&0.928&0.918\\
    \cline{2-5}
    &coarse-grained+negation&0.675&0.911&0.905\\
    \hline
    \multirow{3}{*}{Test}&Fine-grained&0.701&0.829&0.826\\
    \cline{2-5}
    &coarse-grained&0.781&0.935&0.927\\
    \cline{2-5}
    &coarse-grained+negation&0.777&0.927&0.923\\
    \hline
 
    \end{tabular}
 \begin{tablenotes}
 
 \item 
\end{tablenotes}
  \end{threeparttable}
\end{adjustbox}
\end{center}
\end{figure}

\section{Data annotation}
We used Brat\footnote{\url{https://brat.nlplab.org/}}, a web-based text annotation tool \cite{stenetorp2012brat}, to create the annotations. Figures \ref{example_all}, \ref{example2_brat}, \ref{example3_brat} illustrate typical examples, showcasing some of the annotation labels and attributes defined in our annotation guidelines. Since real patient data cannot be publicly shared due to their sensitive nature, these examples are synthetics. We rely on the same examples in the section \ref{erroranalysis} for our error analysis part.
\subsection{Dataset}
We annotated data for a cohort of 200 patients, born on or before 1949, from NHS Lothian\footnote{health board in Scotland} hospitals. 100 individuals are aged between 70 and 79 years and 100 are aged 80 years or over. All patients had to have had more than two recent hospital visits.  For this cohort, DataLoch provided us with a set of de-identified discharge summaries (2,040) belonging to these patients.  Two annotators (Annotator1 and Annotator2) are involved in the annotation process. Annotator1 is a junior doctor with several years of clinical experience and Annotator2 is a geriatrician.

From the 2,040 discharge summaries, 50 documents (24,872 words) were used as a pilot set. This subset was annotated by both annotators, followed by a disagreement meeting to discuss and resolve any inconsistencies. All disagreements were settled through discussion, which also helped to clarify ambiguities and refine the annotation guidelines by adding more examples. The remaining 1,990 documents were used for training and evaluating the model. Among them, 409 documents (190,505 words) were randomly selected as the gold standard test set and annotated by the same two annotators who worked on the pilot set. The remaining 1,581 documents, used for model training (1,265 documents/612,338 words) and validation (319 documents/151,981 words), were annotated by a single annotator (Annotator1). This approach aligns with the methodology of \citet{chen2019detecting}, where one annotator was responsible for annotating the entire corpus, while a second annotator was involved only in annotating a subset to establish the gold standard used for the final evaluation.

\begin{figure*}[t]
\begin{center}
     \begin{adjustbox}{width=1\textwidth}

\begin{threeparttable}
\caption{Evaluation of the annotation (entity level)-- comparison of the performances on 5 models}
   \label{eval_entity_doc}

\begin{tabular}{|p{3cm}|p{4cm}|p{1.3cm}|p{1cm}|p{2cm}|p{1.3cm}|p{1cm}|p{2cm}|}
\hline
    \multirow{2}{*}{Model}&\multirow{2}{*}{Annotation type}&\multicolumn{3}{|c|}{Entity level}&\multicolumn{3}{|c|}{Document level}\\
    \cline{3-8}
    &&Precision&Recall&F1-score&Precision&Recall&F1-score\\
    \hline
    \multirow{3}{*}{Glove}&Fine-grained&0.657&0.605&0.630&0.730&0.582&0.648\\
    \cline{2-8}
    &Coarse-grained&0.828&0.784&0.805&0.902&0.902&0.902\\
    \cline{2-8}
    &Coarse-grained+negation&0.817&0.771&0.794&0.877&0.855&0.866\\
    \hline
    \multirow{3}{*}{BERT\_uncased}&Fine-grained&0.655&0.697&0.675&0.698&0.746&0.721\\
    \cline{2-8}
    &Coarse-grained&0.853&0.860&0.857&\textbf{0.932}&0.946&0.939\\
    \cline{2-8}
    &Coarse-grained+negation&0.600&0.576&0.588&0.715&0.666&0.689\\
    \hline
    \multirow{3}{*}{BERT\_cased}&Fine-grained&0.597&0.661&0.627&0.655&0.707&0.680\\
    \cline{2-8}
    &Coarse-grained&\textbf{0.865}&0.884&\textbf{0.874}&0.927&0.959&\textbf{0.943}\\
    \cline{2-8}
    &Coarse-grained+negation&0.828&0.844&0.836&0.897&0.923&0.910\\
    \hline
    \multirow{3}{*}{BioBERT}&Fine-grained&0.655&0.689&0.671&0.714&0.746&0.730\\
    \cline{2-8}
    &Coarse-grained&0.856&0.885&0.870&0.918&0.959&0.937\\
    \cline{2-8}
    &Coarse-grained+negation&0.854&0.859&0.856&0.908&0.930&0.919\\
    \hline
    \multirow{3}{*}{Bio-Clinical\_BERT}&Fine-grained&0.576&0.586&0.581&0.620&0.572&0.595\\
    \cline{2-8}
    &Coarse-grained&0.840&\textbf{0.887}&0.863&0.908&\textbf{0.961}&0.933\\
    \cline{2-8}
    &Coarse-grained+negation&0.816&0.868&0.841&0.874&0.947&0.909\\
    \hline

    \end{tabular}
 \begin{tablenotes}
 
 \item 
\end{tablenotes}
  \end{threeparttable}
\end{adjustbox}
\end{center}
\end{figure*}
\subsection{Inter-annotator agreement (IAA)}
In this study, we focus on both entity- and document-level annotation. At the entity level, the goal is to identify each tag along with its position (from the example shown in Figure \ref{example_all}, Span[9:10] \textit{confusion → Delirium}, Span[131:133] \textit{femur fracture → Hip\_fracture (In\_hospital)}). Unlike the entity-level annotation, document-level annotation does not consider entity positions or repetitions. For this example, the document-level entities include: \textit{Delirium, Vertebral\_compression\_fracture (History\_of), Urinary\_tract\_infection, etc}. We also consider three levels of annotation granularity: fine-grained, coarse-grained, and coarse-grained with negation. For fine-grained annotation, we retain all tags and attributes, resulting in labels such as \textit{Falls+history\_of+In\_hospital\_events}, allowing us to capture specific contextual information. In the coarse-grained annotation, we focus only on the main entities, such as \textit{Falls}. The coarse-grained annotation with negation includes the primary entities along with a single attribute—\textit{negation}, leading to labels like \textit{Falls+Negation}. Figures \ref{coarse_grained_stat} and \ref{coarse_grained_negation_stat} illustrate the distribution of various entities (coarse-grained and coarse-grained with negation) across the train, development, and test datasets. Due to space constraints, the fine-grained distribution is presented in the appendix \ref{fine_grained_stat_fig}

\subsubsection{Entity and document level agreements}
Table \ref{iaa_entity_doc} presents the entity- and document-level agreement scores for the pilot and test datasets, demonstrating improvements in annotation consistency. We use Bratiaa for calculating the f1-score for the entity level\footnote{\url{https://github.com/kldtz/bratiaa/}}.At the entity level, fine-grained annotations showed moderate agreement (\textit{0.532}) in the pilot corpus, while coarse-grained annotations achieved higher alignment (\textit{0.675}). Refining annotation guidelines improved agreement in the test corpus for both fine-grained (\textit{0.701}) and coarse-grained (\textit{0.781}) annotations, highlighting the impact of task complexity on agreement scores. At the document level, the test corpus consistently outperformed the pilot across all annotation types, with fine-grained (\textit{0.829} vs. \textit{0.720}), coarse-grained (\textit{0.935} vs. \textit{0.928}), and coarse-grained with negation (\textit{0.927} vs. \textit{0.911}) annotations showing improvement. These results confirm the effectiveness of pilot studies in refining annotation guidelines and enhancing annotation consistency, particularly in document-level extraction, which is crucial for clinical applications. More detailled agreements for some entities are presented in appendix \ref{resultsdetail}

\setcounter{figure}{1}
\renewcommand{\figurename}{Figure}

\setcounter{figure}{6}
\renewcommand{\figurename}{Table}
\begin{figure*}[t]
\begin{center}
     \begin{adjustbox}{width=1\textwidth}

\begin{threeparttable}
\caption{Some examples}
   \label{apply_model_example}

\begin{tabular}{|p{2cm}|p{4cm}|p{12.5cm}|p{7.5cm}|}
\hline
    Example&Annotation type&Entity level&Document level\\
   \hline
    \multirow{3}{*}{Example 1}&\multirow{10}{*}{Fine-grained}&\textbf{Span[9:10]}:confusion/Delirium(1.0)&Delirium\\
     \cline{3-3}
    &&\textbf{Span[35:38]}: fracture of L2/Vertebral\_compression\_fracture+ Diagnosis+History\_of(0.8957)&Vertebral\_compression\_fracture+Diagnosis Vertebral\_compression\_fracture+History\\
    \cline{3-3}
    &&\textbf{Span[52:53]}: UTI/Urinary\_tract\_infection(0.9993)&Urinary\_tract\_infection\\
    \cline{3-3}
    &&\textbf{Span[62:63]}: fall/Falls(0.9973)&Falls\\
    \cline{3-3}
    &&\textbf{Span[69:71]}: femur fracture/Hip\_fracture+Diagnosis+History\_of(0.4575)&Hip\_fracture\\
    \cline{3-3}
    &&\textbf{Span[92:94]}: subdural haemorrhage/Intracranial\_haemorrhage(0.7101)&Intracranial\_haemorrhage+Diagnosis, Intracranial\_haemorrhage+History\_of\\
    \cline{3-3}
     &&\textbf{Span[107:108]}: seizure/Seizures(0.9212)&Seizures\\
    \cline{3-3}
    &&\textbf{Span[131:133]}: femur fracture/Hip\_fracture+Diagnosis+History\_of(0.8819)&Constipation\\
    \cline{3-3}
    &&\textbf{Span[149:150]}: constipation(0.9975)&\\
    \cline{3-3}
    &&\textbf{Span[158:159]}: delirium/Delirium(0.9978)&\\
    \cline{2-4}
    &\multirow{10}{*}{Coarse-grained}&\textbf{Span[9:10]}:confusion/Delirium(1.0)&Delirium\\
     \cline{3-3}
    &&\textbf{Span[35:38]}: fracture of L2/Vertebral\_compression\_fracture(0.9998)&Vertebral\_compression\_fracture\\
    \cline{3-3}
    &&\textbf{Span[52:53]}: UTI/Urinary\_tract\_infection(1.0)&Urinary\_tract\_infection\\
    \cline{3-3}
    &&\textbf{Span[62:63]}: fall/Falls(1.0)&Falls\\
    \cline{3-3}
    &&\textbf{Span[69:71]}: femur fracture/Hip\_fracture(0.9999)&Hip\_fracture\\
    \cline{3-3}
    &&\textbf{Span[92:94]}: subdural haemorrhage/Intracranial\_haemorrhage(0.9997)&Intracranial\_haemorrhage\\
    \cline{3-3}
     &&\textbf{Span[107:108]}: seizure/Seizures(0.9997)&Seizures\\
    \cline{3-3}
    &&\textbf{Span[131:133]}: femur fracture/Hip\_fracture(0.8958)&Constipation\\
    \cline{3-3}
    &&\textbf{Span[149:150]}: constipation/Constipation(0.9999)&\\
    \cline{3-3}
    &&\textbf{Span[158:159]}: delirium/Delirium(1.0)&\\
    \cline{2-4}
    (Figure \ref{example_all})&\multirow{10}{*}{Coarse-grained+negation}&\textbf{Span[9:10]}:confusion/Delirium(1.0)&Delirium\\
     \cline{3-3}
    &&\textbf{Span[35:38]}: fracture of L2/Vertebral\_compression\_fracture(0.953)&Vertebral\_compression\_fracture\\
    \cline{3-3}
    &&\textbf{Span[52:53]}: UTI/Urinary\_tract\_infection(1.0)&Urinary\_tract\_infection\\
    \cline{3-3}
    &&\textbf{Span[62:63]}: fall/Falls(1.0)&Falls\\
    \cline{3-3}
    &&\textbf{Span[69:71]}: femur fracture/Hip\_fracture(0.9996)&Hip\_fracture\\
    \cline{3-3}
    &&\textbf{Span[92:94]}: subdural haemorrhage/Intracranial\_haemorrhage(0.9972)&Intracranial\_haemorrhage\\
    \cline{3-3}
     &&\textbf{Span[107:108]}: seizure/Seizures(0.944)&Seizures\\
    \cline{3-3}
    &&\textbf{Span[131:133]}: femur fracture/Hip\_fracture(0.9986)&Constipation\\
    \cline{3-3}
    &&\textbf{Span[149:150]}: constipation/Constipation(0.9965)&\\
    \cline{3-3}
    &&\textbf{Span[158:159]}: delirium/Delirium(0.9999)&\\
    \hline
    
    \end{tabular}
 \begin{tablenotes}
 
 \item 
\end{tablenotes}
  \end{threeparttable}
\end{adjustbox}
\end{center}
\end{figure*}

\section{Evaluation of the annotation on the test corpus}
\subsection{Methodology}
All models were implemented using the FlairNLP framework \cite{akbik2019flair}, following a standard LSTM-CRF architecture. Each model leveraged one of five types of pre-trained embeddings: GloVe \cite{pennington2014glove}, BERT (cased and uncased) \cite{devlin2019bert}, BioBERT \cite{lee2020biobert}, or BioClinicalBERT \cite{alsentzer-etal-2019-publicly}. These embeddings were passed into a bidirectional LSTM \cite{hochreiter1997long} encoder consisting of two recurrent layers with a hidden size of 64 and a dropout rate of 0.16. A Conditional Random Field (CRF) \cite{lafferty2001conditional} layer was used for the final sequence labeling task. The models were trained for 200 epochs using stochastic gradient descent (SGD) with a learning rate of 0.037 and a mini-batch size of 2, constrained by GPU availability within the Trusted Research Environment (TRE). As a result, each model took approximately 36 hours to complete training. Due to the computational limitations and runtime restrictions in the TRE, we did not perform hyperparameter optimisation. Additionally, to support the annotation of discontinuous and overlapping entities, we adapted preprocessing methods from \citet{dai2020effective} to convert Brat annotations into a compatible CoNLL format \footnote{\url{https://universaldependencies.org/format.html}}. Next, we generated three versions of the dataset from each CoNLL file: fine-grained (retaining all transformed entities and attributes), 
coarse-grained (keeping only the main entities) and coarse-grained with negation (including all entities along with the negation attribute). Each of these versions was then used to train models with five different embeddings: Glove , BERT \cite{devlin2018bert} uncased, BERT cased, BioBERT \cite{lee2020biobert} and BioClinicalBERT \cite{alsentzer-etal-2019-publicly}.

\subsection{Entity and document level results}
Table \ref{eval_entity_doc} presents the entity and document level evaluations of the five models used across fine-grained, coarse-grained, and coarse-grained with negation annotations. Coarse-grained annotations consistently outperformed fine-grained annotations across all models, as expected due to fewer classes and more training examples. For the entity level evaluation, BERT (cased) achieved the highest performance for coarse-grained annotations  (\textit{F1 = 0.874}), followed closely by BioBERT (\textit{F1 = 0.870}). When negation attributes were included, these models maintained strong performance, with BERT (cased) scoring \textit{0.836} and BioBERT scoring \textit{0.856}, demonstrating their robustness in handling annotation complexity. Bio-Clinical BERT also performed well for negation detection (\textit{F1 = 0.841}). The results also indicate that document-level extraction benefits significantly from coarse-grained annotations, with models achieving high F1-scores. BERT (cased) performed best (\textit{F1 = 0.943}), surpassing BioBERT (\textit{F1 = 0.937}). These findings highlight the effectiveness of transformer-based models, particularly BERT variants, in leveraging contextual information for document-level tasks. 

Fine-grained annotations posed greater challenges, yielding lower scores across all models. BioBERT and BERT (uncased) performed best for fine-grained annotations, with F1-scores of \textit{0.730} and \textit{0.721}, respectively (for document level). However, GloVe and Bio-Clinical BERT struggled, reflecting the increased complexity and ambiguity of fine-grained annotations. The highest F1-score for fine-grained annotations was \textit{0.675}, achieved by BERT (uncased), followed by BioBERT (\textit{F1 = 0.671}), for the entity-level. GloVe and Bio-Clinical BERT performed the worst, with Bio-Clinical BERT scoring below \textit{0.630}, highlighting the impact of annotation granularity on model performance.

\subsection{Error analysis}
\label{erroranalysis}
 To evaluate the performance of our models, we apply them to some synthetic examples, which were created as part of our annotation guideline development. Table \ref{apply_model_example} displays the different AEs extracted (at both entity and document levels) for each level of granularity we considered for the our presented example in Figure \ref{example_all} (We focus on just one example because of space constraint but four other examples are presented in Appendix \ref{morebratexamples}). For each granularity type, we use the best-performing model: BioBERT for entity-level and BERT\_uncased for document-level in fine-grained analysis, BERT\_cased for both levels in coarse-grained analysis, and BioBERT for both levels in coarse-grained+negation analysis.

 The tagging of this example aligns well with the presented results, as all coarse-grained entities were successfully detected at both the entity and document levels. However, the models exhibited difficulty in identifying the In-hospital and implicit\_mention attributes, as well as in detecting the negation of \textit{delirium}. This can be primarily attributed to the limited number of instances of these categories in the training data. For instance, the dataset contains only six instances of \textit{Seizures\_in-hospital} and nine instances of \textit{Constipation\_in-hospital}, which limits the model’s ability to learn these patterns effectively. Moreover, as the focus of this study is to highlight the dataset and annotation guidelines, we did not explore techniques such as data balancing or adjusting class weights to mitigate the effects of underrepresented categories. Addressing these limitations remains an important objective for future work.

\section{Conclusion}
This study presents a manually annotated dataset for extracting Adverse Events (AEs) from clinical text, specifically focusing on discharge summaries of elderly patients. While our work does not propose a novel method or architecture for language analysis, its primary contribution lies in the development of a high-quality, richly annotated dataset tailored for a critical real-world application: the extraction of adverse events (AEs) from clinical discharge summaries of elderly patients — a population often underrepresented in existing corpora. The novelty of our work resides in the annotation schema, which was carefully designed in collaboration with clinicians to capture clinically meaningful entities and attributes. It supports complex linguistic phenomena such as discontinuous and overlapping entities, as well as a multi-attribute tagging system (e.g., negation, suspected, referral, in-hospital), which are rarely addressed in previous AE-related datasets. Additionally, this resource enables more nuanced model development and benchmarking for clinical NLP, while also surfacing real-world challenges like entity imbalance, fine-grained annotation complexity, and the computational limitations inherent in Trusted Research Environments (TREs).
Beyond fine-grained extraction, our dataset is also compatible with research tasks involving more generic annotation schemes, such as simply labeling the presence or absence of adverse events at the document level — which aligns with existing efforts like the n2c2 corpus (a subset of MIMIC annotated for AEs)

This dataset has also been extended for the detection of geriatric syndromes \cite{guellil2024enhancing} and the social context of patients. Additionally, as future work, we plan to assess cross-domain generalisation by applying our annotation guidelines to the MIMIC-IV dataset \cite{johnson2023mimic}. Inspired by the approach in \cite{dai2024multiade}, which introduced a multi-domain benchmark for AE detection, this evaluation will help determine how well models trained on Dataloch generalize to MIMIC-IV, and vice versa, further advancing AE detection in clinical settings.

\section{Limitations}
Working within a TRE presented several challenges. Strict access policies and security rules, while necessary for data protection, made it difficult to train models in real-time. Additionally, limited GPU resources slowed down model development, especially for deep-learning models that require high computational power. These restrictions forced us to carefully choose and optimize models to make the best use of available resources. We are also constrained by the variety of language models we are allowed to use. State-of-the-art large language models, such as GPT-4 or Mistral, are not provided for use in DataLoch's TRE due to patient privacy concerns, which rules out the use of SOTA LLMs

One of the main challenges in this study was the imbalance of AE information in the dataset. This can be seen in the appendix, where the detailed results for each entity are presented. We noticed that the best-performing entities were also the most frequent in the corpus, such as falls and delirium. In contrast, rare entities had lower performance, which could have been improved by balancing techniques or assigning different weights to each class. However, due to limited resources, we were unable to test these approaches. Additionally, we did not optimize the hyperparameters. FlairNLP uses Hyperopt for optimization, but running just 50 iterations on the entire dataset would have taken more than three weeks. Since the TRE instance restarts randomly, it often cancels long-running tasks, making hyperparameter tuning impractical in this setting.
In summary, working with real patient data offers many advantages, but access to this type of data is restricted to Trusted Research Environments (TREs) due to privacy and security concerns. As a result, applying state-of-the-art techniques within these environments can be slower and more time-consuming.

\section*{Acknowledgment}
This research has been assisted by the DataLoch service (reference: DL\_2021\_009) and received favourable ethical opinion under DataLoch's delegated Research Ethics authority (Reference: 17/NS/0027). DataLoch enables access to de-identified extracts of health care data from the South-East Scotland region to approved applicants: \url{dataloch.org}.

For MIMIC IV, the collection of patient information and creation of the research resource was reviewed by the Institutional Review Board at the Beth Israel Deaconess Medical Center, who granted a waiver of informed consent and approved the data sharing initiative.

This study/project (AIM-CISC) was funded by the National Institute for Health Research (NIHR). Artificial Intelligence and Multimorbidity: Clustering in Individuals, Space and Clinical Context (AIM-CISC) grant NIHR202639. The views expressed are those of the author(s) and not necessarily those of the NIHR or the Department of Health and Social Care.

This research was also funded by the Legal \& General Group (research grant to establish the independent Advanced Care Research Centre at the University of Edinburgh). The funder had no role in conduct of the study, interpretation or the decision to submit for publication. The views expressed are those of the authors and not necessarily those of Legal \& General.

As this work is being extended to MIMIC IV, the authors would like to express their sincere gratitude to all the annotators involved in this annotation, including Scott Prentice, Mike Holder, Rose Penfold and Lauren Ng. Finally, the authors would like to warmly thank Richard Tobin (School of Informatics) for his amazing help in setting up Brat. They would also like to thank the ACRC management team for their valuable support.

\bibliography{custom}
\section*{Appendices}
\appendix
\section{Table comparing the proposed studies to the proposed work}
\label{table_related}
Table \ref{relatedwork} provides a comprehensive summary of prior studies included in the related work section. For each study, we specify its classification, the methodological approach employed, the datasets utilized, the number of annotators involved in the annotation process, the models and machine learning algorithms applied, and whether the authors provide annotation guidelines for manually annotated corpora. Additionally, we examine whether the studies consider discontinuous and overlapping entities. Furthermore, we present the same elements for our proposed work to underscore our contribution and facilitate a clear comparison between existing research and the novel aspects of our study.

\setcounter{figure}{0}
\renewcommand{\figurename}{Table}
\begin{figure*}[h]
\begin{center}
     \begin{adjustbox}{width=1\textwidth}

\begin{threeparttable}
\caption{Related work summary}
   \label{relatedwork}

\begin{tabular}{|p{2.5cm}|p{2.2cm}|p{2cm}|p{3cm}|p{3cm}|p{2cm}|p{1.5cm}|p{3cm}|p{2.1cm}|p{1.8cm}|p{2cm}|p{3cm}|p{3cm}|}
\hline
    Work&Category&Approach&Datasets / lexicons&Sources&Annotation guideline&Number annotators&Entities&Discontinuous entities&Overlapping&IAA used&Model embedding/ontology&ML algorithm\\
   \hline
    \cite{tafti2017adverse}&Classification&Binary classification (sentence level)&Biomedical articles + social media data &Pubmed + MedHelp + WenMD + patientinfo&No&3&AEs/No-AEs&No&No&Kappa&Word2vec&BigNN + SVM, decision tree+ NB\\
    \hline
    \cite{ellenius2017detecting}&Classification&Binary classification&Medical product tweets / VigiBase medical event dictionary&Twitter&No&-&suspected AEs&No&No&-&MedDRA&LR\\
    \hline
    \cite{chen2019detecting}&Classification&Binary classification&secure messages threads (patients with diabetes and healthcare provider)&U.S. Department of Veterans Affairs system&yes (simple one)&2 (one the whole dataset and the second for a subset)&Hypoglycemia / No-Hypoglycemia&No&No&Kappa&Word2vec&LR / RF / SVM with class wighting (SMOTE)\\
    \hline
    \cite{ujiie2020identification}&Classification&Binary classification&Japanese medical articles&Japanese pharmaceutical company&No&2&AE, NoAE&No&No&Kappa&bag of words, standard disease, context word tokens, etc&logistic regression\\
    \hline
    \cite{shen2021automatic}&Classification + extraction&IBM Watson Natural Language Understanding API&Electronic Medicines Compendium (EMC)&European Medicines Agency&No&2&ADE termes&No&No&-&-&Rule-based + IBM Watson\\
    \hline
    \cite{yang2019madex}&Classification +extraction + relation extraction&Binary classification +NER&MADE dataset&EHR from the University of Massachusetts Medical School&Yes&-&AE +drug +dose +frequency + route +duration&Yes&Yes&-&word + character level embedding&SVM + LSTM-CRF\\
    \hline
    \cite{kim2018machine}&Classification&multiple class classification&Korea Adverse Event Reporting System (KAERS) database&Korea Institute of Drug Safety and Risk Management (KIDS)&No&-&Certain, Probable, Possible, Unlikely, Unclassified, Unclassifiable&No&Yes&-&TF-IDF&NB +Laplace smoothing\\
    \hline
    \cite{portelli2021improving}&Extraction&NER&SMM4H +CADEC&Twitter + AskaPatient&No&-&AEs&No&No&-&BERT +SpanBERT + BioBERT&CRF\\
    \hline
    \cite{wu2021chinese}&Extraction&NER&AE reports&Jiangsu Province AE Monitoring Center  +Drum Tower Hospital&No&-&Drug, reason, AE&No&No&-&BERT&Bi-LSTM +CRF\\
    \hline
    \cite{emadzadeh2018hybrid}&Normalisation&Entity linking&corpus of tweets +UMLS&Twitter +the U.S. National Library of Medicine (NLM)&No&3&AE, Indication(symptoms), drugs&No&No&agreement based on a third annotator&UMLS&Rule-based matching +LSA +HSA +regression model\\
    \hline
    \cite{combi2018normalizing}&Normalisation&Mapping&spontaneous reports&pharmacovigilance databases&No&-&AEs&No&No&-&MEDDRA&String similarity algorithms +contextual analysis\\
    \hline
    \cite{munkhdalai2018clinical}&Extraction&NER +relation extraction&EHR narratives&clinical setting&No&-&AE, drugs, indication, severity&No&No&F1 score (document level)&one-hot encodding +n-grams&SVM + RNN + LSTM + Supervised rule induction\\
    \hline
    \cite{florez2019deep}&Extraction +relation extraction&NER&n2c2&MIMICIII&No&-&Drugs, AEs, Attributes (dosage, route, duration)&No&No&-&DeBERTa, MeDeBERTa, RoBERTa&Rule-based, NN\\
    \hline
    \cite{lavertu2021quantifying}&ADE severity&semi-supervised method&FAERS +Gottlieb severity \cite{gottlieb2015ranking}&openFDA website&No&-&severity of AEs&No&No&-&RedMed +MedVRA&Lexical network +label propagation algorithms\\
    \hline
    \cite{zhang2021identifying}&Extraction&NER&Twimed-pubmed&Twitter +Pubmed&No&-&AE, drugs&No&No&-&Word2vec&CharCNN\\
    \hline
    \cite{karimi2015cadec}&Resource construction +normalisation&Manual annotation&CADEC&AskaPatient&Yes&4&Drug, ADR, Disease, Symptom, Finding&Yes&Yes&customised metric&SNOMED CT, MedDRA&-\\
    \hline
    \cite{yahya2022automatic}&Extraction&lexicon-based approach&Patient review +Consumer Health Vocabulary + SIDER&Askapatient + WebMD&No&-&AE for aniti-epileptic drugs&No&No&-&No&Lexicon-based approach\\
    \hline
    \cite{d2023biodex}&Extraction + severity of events&dataset construction + few shot learning&BioDEX&PubMed articles + Drug Safety Reports&No&-&AE, drugs, seriousness, patient gender&No&No&F1 score&text-davinci-002, gpt-3.5, gpt-4, FLAN-T5-Large, FLAN-T5-XL&-\\
    \hline
    \cite{li2024improving}&Extraction&NER&reports + posts&VAERS + Twitter + Reddit&No&-&Vaccine, Shot, AE&No&No&-&BioBERT, GPT-2, GPT-3.5, GPT-4, Llma-2 7b, Llma-2 13b&RNN\\
    \hline
    Our work&Extraction&NER (Token + document level)&discharge summaries&Dataloch&Yes&2 (one the whole data and the other for a subset)&AEs (falls, delirium, hip-fracture, intracranial hemorrhage, lower tract infection, etc.&Yes&Yes&F1-score (token level + document level) and Kappa (document level)&Glove, BERT uncased, BERT cased, BioBERT, BIO-Clinical BERT&CRF + LSTM (FlairNLP)\\
    \hline
    
    \end{tabular}
 \begin{tablenotes}
 
 \item 
\end{tablenotes}
  \end{threeparttable}
\end{adjustbox}
\end{center}
\end{figure*}

\section{Detailed annotation guideline provided to the annotators}
\label{detailedguideline}
\subsection{Introduction}
The purpose of this work is to extract Adverse Events (AE) in elderly patients' electronic health records (discharge summaries) to be used as a novel source of data for health and social care research. In order to automate this process, we rely on Natural Language Processing (NLP), a sub-field of artificial intelligence which makes use of machine learning to analyse human language. This requires the creation of a manually annotated dataset of clinical free-text on which the NLP model can be trained on. In this annotated corpus, the elements of interest (AE) are manually highlighted and given a label. The aim of this document is to provide a guideline for this manual annotation. 

Note that all examples used in this document are created by clinicians in our team for the purpose of this document, they are not extracted from clinical data.

\subsection{Annotation categories}
Adverse Events (AE) can be defined as harmful events or undesired harmful effects caused by healthcare management rather than the patient's underlying disease. They have a wide range of severity and can result in prolonged hospital stay, permanent disability or contribute to death \cite{rawal2019identification,baker2004canadian,vincent2001adverse}. Different AE are referenced in the literature including major osteoporotic fractures, intracranial haemorrhage and constipation. In this work, we focus on 14 AEs: falls, delirium, pressure injury, hip fractures, wrist fractures, proximal humeral fractures, vertebral compression fractures, intracranial haemorrhage, upper gastrointestinal haemorrhage, lower gastrointestinal haemorrhage, lower respiratory tract infection, urinary tract infection, constipation and seizures.

We present in the following part, the list of AEs we wish to extract with a short definition for each of them: 

\begin{enumerate}
    \item Falls: \textit{are an events which result in a person coming to rest inadvertently on the ground or floor or other lower level}\footnote{\url{https://www.who.int/news-room/fact-sheets/detail/falls}}. \textit{Falls among the elderly, including same-level falls, are a common source of both high injury severity and mortality, much more so than in younger patients} \cite{sterling2001geriatric}.
    \item Delirium: \textit{is an acute state of brain failure marked by sudden onset of confusion, a fluctuating course, inattention, and often an abnormal level of consciousness} \cite{mattison2020delirium}. \textit{It is generally reversible and can occur in response to a wide range of precipitants including acute illness, surgery, trauma or drugs, though a clear precipitant is not identifiable in around 10\% of cases} \cite{maclullich2009delirium}.
    \item Pressure injury: \textit{is damage to the skin and the deeper layers of tissue under the skin caused by prolonged pressure to an area of the body. It is more likely to happen if a person has to stay in a bed or chair for a long time.\footnote{\url{https://www.nice.org.uk/guidance/cg179/ifp/chapter/what-is-a-pressure-ulcer}} This term includes any soft tissue injury, even without ulceration} \cite{edsberg2016revised}\textit{ and it is also known as "decubitus ulcer", "bedsore" or "pressure sore".}
    \item Hip fractures: \textit{Proximal femoral fractures (including femoral neck, intertrochanteric and subtrochanteric fractures) are the most common reason for admission to an acute orthopaedic ward and are associated with excess mortality up to 36\% in the first year} \cite{abrahamsen2009excess}. \textit{Over 35\% of patients will not return to their pre-fracture functional status after 12 months} \cite{magaziner1990predictors} \textit{and 18\% of patients require a more care-intensive environment within the following year}\cite{schurch1996prospective}\textit{. This definition excludes periprosthetic fractures, isolated fractures of the greater trochanter, the pubic rami or the acetabulum} \cite{hall2022impact}.
    \item Wrist fractures: \textit{These consist of a fracture of the distal end of the radius, the distal end of the ulna or both, and over 90\% of them involve a fall} \cite{nevitt1993type}. \textit{They are the most common amongst women until the age of 75, after which their frequency is surpassed by hip fractures} \cite{cummings1985epidemiology}. \textit{Furthermore, while wrist fractures are rarely fatal and cause much less disability than hip fractures, a functional decline can be observed at 6 months post-fracture in 33\% of patients over the age of 65} \cite{vergara2016wrist}.
    \item Proximal humeral fractures: \textit{They are the third most frequent fracture in the elderly following hip and wrist. While the long-term functional outcome is satisfactory in 80\% of patients after a simple humeral fracture without displacement, displaced proximal humeral fractures may require lengthy hospitalisation and generally lead to long-term functional deficit} \cite{lee2002risk}\textit{. They include humeral fractures of the anatomical neck, the great tuberosity, the proximal end, the surgical neck and the upper epiphysis.}
    \item Vertebral compression fractures: \textit{Usually diagnosed on radiography, a vertebral fracture is confirmed when there is an approximate 20\% loss in vertebral body height relative to normal looking adjacent vertebra} \cite{griffith2015identifying}. \textit{Approximately 25\% of all postmenopausal women in the US get a compression fracture during their lifetime. Osteoporosis is the most common aetiology for these fractures and they can occur during trivial events such as lifting a light object, a vigorous cough or sneeze or turning in bed. Although vertebral compression fractures rarely require hospital admission, they have the potential to cause significant disability and morbidity, often leading to incapacitating back pain for many months} \cite{alexandru2012evaluation}.
    \item Intracranial haemorrhage: \textit{This term refers to any bleeding within the intracranial vault, including the brain parenchyma and surrounding meningeal spaces. This definition incorporates four main diagnoses based on anatomical location: epidural haematomas, subdural haematomas, subarachnoid haemorrhages and intracerebral haemorrhages (including parenchymal contusions). Intracranial haemorrhage may be spontaneous, precipitated by an underlying vascular malformation, induced by trauma, or related to therapeutic anticoagulation. Subdural and epudural haematomas are usually traumatic injuries} \cite{naidech2011intracranial}\textit{. Subdural haematomas are much more common, they occur in about 30\% of severe head injuries and the underlying brain damage is usually much more severe than with epidural heamatomas} \cite{tallon2008epidemiology}\textit{. Non-traumatic subarachnoid haemorrhages are most commonly due to the rupture of an intracranial aneurysm and for intracerebral haemorrhages, hypertension is the most important risk factor} \cite{qureshi2001spontaneous}\textit{. Most patients who survive intracranial haemorrhage have a markedly lower quality of life than the general population. The 30-day mortality rate of spontaneous intracerebral haemorrhages ranges from 35\% to 52\% with only 20\% of survivors expected to have full functional recovery at 6 months} \cite{caceres2012intracranial}.
    \item Upper gastrointestinal haemorrhage: \textit{It is a common medical emergency worldwide and refers to bleeding from the oesophagus, stomach, or duodenum. Patients present with haematemesis or melaena, although haematochezia can occur in the context of a major bleed and is typically associated with haemodynamic instability} \cite{stanley2019management}\textit{. Peptic ulcer bleeding is the most common cause of upper gastrointestinal bleeding, responsible for about 50\% of all cases, followed by oesophagitis and erosive disease. Variceal bleeding is less frequent in the general population, however, it has been found to be the cause of bleeding in cirrhotic patients in 50–60\%} \cite{van2008epidemiology}.
    \item Lower gastrointestinal haemorrhage: \textit{Haematochezia is the typical symptom of lower GI bleed, although as mentioned previously, some severe upper GI haemorrhages can also present with red blood in the stools. Similarly, while melaena typically indicates bleeding from a foregut location, it may result from bleeding from the small intestine or the right colon, particularly with slow GI bleeding or slow GI transit} \cite{ghassemi2013lower}\textit{. The most common causes of lower gastrointestinal bleeding are diverticulosis, angiodysplasia, haemorrhoids, and ischaemic colitis} \cite{farrell2005management}.
    
    \item Lower respiratory tract infection (LRTI): \textit{These include any bacterial, viral or fungal infection of the respiratory tree and will usually manifest as a pneumonia, bronchitis or lung abscess. LRTI developed in hospital (healthcare-associated pneumonias or hospital acquired pneumonias) are usually caused by bacteria. In these cases, attributable mortality rates of 20 to 33\% have been reported} \cite{tablan2003guidelines,american2005guidelines}\textit{ and its presence increases hospital stay by an average of 7 to 9 days.}
    \item Urinary tract infection (UTI): \textit{These include infections of the bladder (cystitis), urethra (urethritis) or kidneys (pyelonephritis) and are most frequently caused by bacteria. When acquired in hospital, they are the most frequent healthcare-associated infections and account for more than 40\% of all healthcare-associated infections in a general hospital} \cite{wagenlehner2012epidemiology} \textit{. The vast majority of these are associated with the presence of an indwelling urinary catheter} \cite{lo2014strategies}.
    
    \item Constipation: \textit{Constipation is used to describe a variety of symptoms, including hard stools, excessive straining or infrequent bowel movements. Many disorders can cause constipation including neurologic disease (e.g. spinal cord injury, Parkinson's disease or multiple sclerosis), rectoanal problems (e.g. anal strictures, proctitis), iatrogenic conditions (e.g. drugs or previous surgeries), endocrine and metabolic diseases (e.g. diabetes or hypothyroidism), lifestyle factors such as lack of mobility and dietary factors, namely low residue diets} \cite{camilleri2017chronic,schiller2001therapy}\textit{. A number of studies have shown that constipation is significantly associated with reductions in elderly patients’ quality of life and functional status} \cite{norton2006constipation,o1995bowel,donald1985study,whitehead1989constipation}.
   
    \item Seizures: \textit{Seizures involve sudden, temporary, bursts of electrical activity in the brain that change or disrupt the way messages are sent between brain cells. These electrical bursts can cause involuntary changes in body movement or function, sensation, behaviour or awareness}\footnote{\url{https://www.epilepsy.com/what-is-epilepsy/understanding-seizures}}\textit{. They can happen in the context of epilepsy or, be provoked by an acute medical illness. In the case of provoked seizures, the main causes are trauma, central nervous system infections, space-occupying lesions, cerebrovascular accidents, metabolic disorders, and drugs} \cite{kaur2018adult}. 

\end{enumerate}

\subsection{General instructions}

\subsubsection{Background}

Named entity recognition is the NLP method we are employing to extract structured information from our unstructured data. This task aims to precisely locate and classify the AE within the clinical free text documents by assigning tokens with a given label.

Tokens are units of text, usually words or abbreviations and are separated by punctuation marks such as spaces, commas, full stops, slashs, hyphen, brackets or numbers.

Labels are used in many different domains, the most common example would be their use for de-identification: automatically detecting and redacting the names of persons, organisations or locations from sensitive documents. For example: 
\begin{itemize}
    \item \hl{Mark Elliot Zuckerberg} is the founder of \hl{Facebook}. He lives in \hl{LA}.
\end{itemize}

In this example, the label \textit{Person} should be associated with the token "Mark Elliot Zuckerberg", the label \textit{Organisation} should be associated with the token "Facebook" and the label \textit{Location} should be associated with the token "LA". This demonstrates that an entity could be extracted from a single word (i.e. Facebook), a set of words (i.e. Mark Elliot Zuckerberg) or an acronym (i.e. LA). 

Annotators must be careful, however, when selecting multiple tokens, to always aim for as few words as possible while retaining the meaning of the label they are tagging. This rigorous process is essential to the training of the language model and will also allow a more uniform annotation between different clinicians.

The de-identification example given above is rather straightforward with a single word or a short string of adjacent words matching one label. Medical text is often more complex: terms can be fragmented and non-adjacent and some entities can overlap, meaning that one word can describe more than one label. Discontinuous entities and overlapping entities are detailed further below.

\subsubsection{Discontinuous entities:} 
Medical texts often contain distant entities which need to be annotated together to make sense of the label which is being recognised. Discontinuous entities, also named distant entities or fragmented entities, are often used for this purpose \cite{huang2023t}. For example:

\begin{itemize}
    \item “The \hl{wrist} X-ray performed in A\&E confirms the presence of a \hl{fracture}” - These two distant tokens should be annotated as one discontinuous entity with the label \textit{Wrist fracture}
\end{itemize}

When deciding to extract discontinuous entities, the annotator should aim to select terms which are as close as possible to each other in the text while retaining the meaning of the label, as shown in the examples below:

\begin{itemize}
    \item "The X-ray of her left femur showed degenerative changes of her \hl{hip} joint and a displaced extracapsular \hl{fracture}" - The highlighted text is given the label \textit{proximal femoral fracture} (or \textit{hip fracture}). In this example, rather than using the term "femur", the annotator should aim to reduce the distance between the two elements of the text which form the fragmented entity and select "hip" which is closer to the word "fracture".
    \item “The wrist X-ray performed in A\&E confirms the presence of a \hl{fracture} of the \hl{radius}” - The highlighted text is given the label \textit{wrist fracture} and instead of selecting the word "wrist", the annotator should use the term "radius" in their annotation of this fragmented entity because it is closer to the word "fracture" while retaining the meaning of the label.
\end{itemize}

This tool should not be used when annotating two different, isolated symptoms which are associated with the same AE. For example:

\begin{itemize}
    \item “She was \hl{confused} on the ward” – Is an appropriate token to select for the label \textit{Delirium}
    \item “She was \hl{agitated} on the ward” – Is also an appropriate token to select for the label \textit{Delirium}
    \item “She was \hl{confused} and \hl{agitated} on the ward” - since both of these symptoms, in isolation, are sufficient to detect delirium, the tokens highlighted in this example must be annotated as two separate entities with the label \textit{Delirium}. The discontinuous entity tool should not be used in this case.
\end{itemize}

\subsubsection{Overlapping entities:} Medical texts also tend to contain overlapping entities which require the annotator to highlight the same element of text twice and this can be used in combination with the distant entities \cite{huang2023t}. For example:

\begin{itemize}
    \item “When she fell, she \hl{broke} her \hl{wrist} and her \hl{hip}” - This can be annotated with “broke” + “wrist” for the label \textit{Wrist fracture} and “broke” + “hip” for the label \textit{hip fracture}.

\end{itemize}

\subsubsection{Supplementary instructions}

\paragraph{Tokens} The annotator should select tokens in their entirety, and never splitting them, even in the case of abbreviations and acronyms. For example:
\begin{itemize}
    \item "She was admitted with a catheter-associated \hl{UTI} and her catheter was replaced with antibiotic cover"
    \item "She was admitted with a \hl{CAUTI} and her catheter was replaced with antibiotic cover" - These two examples demonstrate that although we do not need to include "catheter-associated" in our selection of tokens when labelling \textit{Urinary tract infection}, we must include "CA" in "CAUTI" because the latter forms one, single token.
\end{itemize}

Only the necessary tokens should be highlighted. Pronouns, articles, and numbers should be excluded from the selection unless they are crucial to identifying AE. For example: 
\begin{itemize}
    \item "Xray confirms a \hl{fracture} of the right \hl{wrist}" - We do not want to include the side of the fracture (right or left) in our selection nor prepositions ("of") and articles ("the"), therefore only the terms "fracture" and "wrist" should be selected. This example makes use of the discontinuous entity tool.
\end{itemize}
We also recommend that annotators also avoid including punctuation in their selection wherever possible. 
\begin{itemize}
    \item "He had no \hl{melaena}/\hl{haematochezia}" - These two tokens can be isolated since they are separated by a slash (/)
\end{itemize}
We acknowledge that on occasions, the selection of punctuation cannot be avoided, for example:
\begin{itemize}
    \item "The hip Xray confirmed a \hl{femoral \#}" - the hash sign "\#" is often used by clinicians to mean "fracture"
\end{itemize}

\paragraph{Repetition:} If an AE appears multiple times within a document, every instance of it should be annotated. For example: 
\begin{itemize}
    \item "She \hl{fell} at home. This \hl{fall} was likely caused by postural hypertension. She has been struggling with her gait since the \hl{fall} because she lost confidence in her balance."
\end{itemize}

\paragraph{Number of fragments:} When deciding to extract discontinuous entities, the annotator must restrict the use of the discontinuous tool to only two fragments. For example:
\begin{itemize}
    \item "She \hl{fractured} several bones in January including two ribs and the proximal end of her left \hl{femur}.
\end{itemize}
Although our label \textit{Hip fracture} specifically requires the annotator to extract fractures of the proximal end of the femur (as opposed to the distal end or mid-shaft fractures), we acknowledge that it would be impossible for them to extract the terms "femur", "fracture" and "proximal end" without either selecting a large number of unwanted tokens or using a third fragment with the discontinuous tool. For this reason, most entities will be defined in this document through two core elements which must be present in the selection of text. For example:
\begin{itemize}
    \item \textit{Lower respiratory tract infection} and \textit{Urinary tract infection} must include a reference to the infection itself ("sepsis", "infective", "septic") and its source ("lung", "chest", "urinary", "bladder") although some words in isolation contain both pieces of information ("pneumonia", "pyelonephritis")
    \item All major osteoporotic fractures must include a term associated with the fracture itself ("broke", "fractured") and their location ("wrist", "radius", "humeral", "vertebra")
\end{itemize}

\subsection{Annotation labels}
The AE category has a set of 14 annotation labels which describe the information we are interested in capturing: hip fracture, proximal humeral fracture, wrist fracture, vertebral compression fracture, intracranial haemorrhage, upper gastrointestinal haemorrhage, lower gastrointestinal haemorrhage, urinary tract infection, lower respiratory tract infection, constipation and seizures. The following examples briefly demonstrate the annotation of these labels:

\begin{enumerate}
     \item "Three \hl{falls} in last yr" \textit{(Falls)}
      \item "On admission, she was acutely \hl{confused}" \textit{(Delirium)}
      \item "Nursing staff has noted a \hl{decubitus ulcer} on the right heel" \textit{Pressure injury}
     \item "Admitted with a left intertrochanteric neck of \hl{femur fracture}" \textit{(Hip fracture)}

    \item "Distal \hl{radial fracture} shown on the X-ray." \textit{(Wrist fracture)} 

    \item "The pain in his shoulder is the result of a \hl{humeral fracture}" \textit{(Proximal humeral fracture)} 

    \item "She has several known osteoporotic \hl{vertebral fractures}" \textit{(Vertebral compression fracture)}

    \item "Right parietal \hl{subdural haematoma} visible on CT" \textit{(Intracranial haemorrhage)}

    \item "He went to see his GP the day before admission with abdominal pain and 3 episodes of \hl{melaena}." \textit{(Upper gastrointestinal haemorrhage)}

    \item "She reported some \hl{PR bleed}" \textit{(Lower gastrointestinal Haemorrhage)}

    \item "She was recently treated for \hl{pneumonia} by her GP" \textit{(Lower respiratory tract infection)}

    \item "During her stay, she described a new burning sensation when passing urine. Ciprofloxacin was started for \hl{UTI}." \textit{(Urinary tract infection)}

    \item "Patient had \hl{difficulties voiding} despite laxatives. This was resolved with an enema." \textit{(Constipation)}

    \item "Found by the nurse with \hl{generalised tonic-clonic episode}" \textit{(Seizure)}
    
\end{enumerate}

\subsection{Annotation attributes}
Annotation attributes are additional features or characteristics associated with the tag. For example, the tag \textit{Person} could have the attribute age, gender or profession. For the tag \textit{Organisation}, it could be the type of organisation, its number of employees, etc. 

For this project, we use four main attributes: Negation, Diagnosis, in-hospital events and low confidence. 

\subsubsection{Negation:}
Negation can be formulated explicitly with "no" or "not" but can also appear as acronyms (e.g. NAD, meaning nothing abnormal discovered) or be inferred from the text. The annotator should, when possible, not include the negation word(s) in the selection of text but simply highlight the words which reference the AE and add the negation attribute to the label. The following examples highlight how negation can be captured:
\begin{itemize}
    \item "She slipped but did not \hl{fall}." - This should be annotated \textit{Falls} with the negation attribute.
    \item "CT scan found no evidence of \hl{intracranial bleeding}." - This requires the label \textit{Intracranial haemorrhage} and the negation attribute.
\end{itemize}

\subsubsection{Low Confidence}
The \textit{Low confidence} should be selected when the annotator is uncertain of their choice. There are two main situations we recommend the use of this attribute:
\begin{itemize}
    \item If there is insufficient information in the document (see examples below)
    \item If the annotator is faced with an unusual description of events which is, according to the writer of the letter, to be associated with AE. 
\end{itemize}
In the second situation, the purpose of the \textit{Low confidence} attribute is to provide some nuance to the final annotated document and indicate to the model that the text highlighted is unlikely to be associated with this AE, again, if in a different context. For example:
\begin{itemize}
    \item "Her husband said that she had \hl{broken} her \hl{leg} about 6 months ago after a fall" - This should be annotated with the discontinuous entity tool and labeled \textit{Hip fracture}. The \textit{Low confidence} attribute should be added since we cannot be certain from this sentence alone that it was her proximal femur which the patient broke.
    \item "She was admitted with delirium, was confused in the ambulance and had visual \hl{hallucinations} at home. Her daughter said she was describing children running around but they were by themselves in the house. The delirium resolved with rehydration." - Hallucinations are rarely present in delirium but it seems to be described by the author of the letter as such. It should therefore be annotated with the label \textit{Delirium} and the \textit{Low Confidence} attribute.
\end{itemize}

\subsubsection{Diagnosis}
This attribute allows the annotator to further qualify the annotated text when the AE is not directly present in the patient's admission, attendance or report. For this purpose, four tags are available:
\begin{itemize}
    \item \textit{History of}, when selected, qualifies AE which has happened in the past. This can be mentioned in the patient's past medical history or something which happened before the admission or attendance and is not related to the current clinical presentation. 
    \item \textit{Suspected}, qualifies when the writer of the clinical document suspects the patient has a GS or AE (note this is not the suspicion of the annotator - if the annotator is unsure about an AE to be annotated, nuance can be added with the attribute \textit{Low confidence} instead (see previous paragraph)). This attribute can also be used in annotating screening tests which require further investigations to confirm a diagnosis.
    \item \textit{Referral} qualifies when a patient is being referred to a different healthcare professional for the diagnosis of their AE.
    \item \textit{Implicit mention} qualifies an AE which is mentioned in the text but does not apply to the patient. This includes descriptions of the patient's family history or diagnosis given to their spouse. It also should be used when the text details hypothetical situations ("if" statements) or risks of an AE happening (e.g. risk of falls).
    \item Finally, if the \textit{Diagnosis} field is left empty, this signifies that the AE has been diagnosed, is present during the admission or attendance or that it is the reason for the admission or attendance.
\end{itemize}
The examples below illustrate each tag:
\begin{itemize}
    \item "PMH: T2DM, Parkinson's disease, \hl{subdural haemorrhage}." - In this example, the "subdural haemorrhage" has been described in the past medical history (PMH) section of the letter and should be annotated with the label \textit{Intracranial haemorrhage} and the attribute \textit{History of} for the \textit{Diagnosis}.
    \item "His haemoglobin remained stable and was referred to the GI team to discuss outpatient investigations of this haematemesis"- In this example, the \hl{haematemesis} should be annotated as \textit{Upper gastrointestinal haemorrhage} with \textit{referral} attribute.
    \item "CT Head showed acute on chronic \hl{subdural haematoma}." - In this case, there is a historical feature (chronic) as well as a recent event (acute) of the intracranial bleed. The highlighted text should therefore be annotated twice with the same label (\textit{Intracranial haemorrhage}), one with the attribute \textit{History of} and one without any further attribute.
    \item "This patient was admitted following a GP visit who suspects she might have a \hl{wrist fracture}." - Here the wrist fracture is \textit{Suspected} and should be annotated with the matching \textit{Diagnosis} attribute.
    \item "The patchy opacification in the left base may represent an \hl{infective} process or \hl{pulmonary} oedema." - This is an example of a radiology interpretation which maintains a level of doubt onto the final diagnosis. We recommend annotating this as one discontinuous entity of \textit{Lower respiratory tract infection} with the \textit{Suspected} attribute
    \item "The community occupational therapy team will review the patient's house after his discharge to evaluate and minimise the risk of \hl{falls}." - This is a hypothetical mention of the GS \textit{Falls} (with "risk of"). It should therefore be annotated as \textit{Falls} with the attribute \textit{Implicit mention}.
    \item "He was given laxatives for prophylaxis of \hl{constipation}" - The term prophylaxis means the patient was given laxatives as a preventative measure, therefore, they did not experience constipation and it should be annotated with the attribute \textit{Implicit mention}.
    \item "He was admitted with a \hl{fall} and developed \hl{delirium} during his admission due to a \hl{urinary tract infection}." - All three of these diagnoses have happened in direct relation to the admission and therefore should be annotated without any additional \textit{Diagnosis} attribute.
    \hl{Fall} last night and was brought to the hospital today after a long lie." - Although the event happened in the past, it is directly related to the current admission therefore the fall should be annotated without any \textit{diagnosis} attribute.    
\end{itemize}

Some letters might have a template to help clinicians write quicker by simply answering propositions. If elements of the template which contain an AE are not applicable to the patient, they should be annotated with the \textit{Implicit mention}. For example: 
\begin{itemize}
    \item "Type and duration of DVT prophylaxis (only for \hl{hip fracture}): Not applicable" - \textit{Implicit mention}
    \item "Type of \hl{Fall}:  / Site of the fracture:  / Operation date:" - \textit{Implicit\ mention} 
    
\end{itemize}
Clinical documents often contain mentions of previous adverse drug reactions which name the drug and the exact adverse event which occurred (e.g. drug-induced delirium). These should be annotated with the attribute \textit{History of} as shown in the example below:
\begin{itemize}
    \item "Allergies and adverse reactions: Bisoprolol (dizziness), Metformin (\hl{constipation})" - This means the patient has experienced constipation because of the medication Metformin in the past, therefore it should be annotated with the attribute \textit{History of}.
\end{itemize}

\subsubsection{In-hospital event}
This attribute allows the annotator to identify AE for which the onset took place in the hospital. For example, if a patient has a fall on the ward after surgery or if they were admitted with no symptom of delirium but developed acute confusion during their stay. 

This attribute aims to capture the location of the event rather than the location of the diagnosis. This is important to keep in mind since some entities are diagnosed during the admission but their onset will have taken place before the patient came to the hospital. For example, if a patient had a fall two days prior and is admitted and treated for an infection. On day 3 they reported pain in their wrist and the X-ray of their arm shows a fracture but clearly, this adverse event is associated with their fall at home 5 days ago, therefore the attribute \textit{In-hospital event} should not be selected. We present further examples in the following part:
\begin{itemize}
     \item "New cough and wheeze during admission. Treated as hospital-acquired \hl{pneumonia}" - In this case, "pneumonia" should be tagged with the label \textit{Lower respiratory tract infection} and the attribute \textit{In-hospital event} should be selected.
    \item "During her post-operative recovery, she described a new burning sensation when passing urine. CRP was raised, urine dip was positive for blood and leukocytes and Ciprofloxacin was started for \hl{UTI}." - This should be tagged as with the label \textit{Urinary tract infection} and the attribute \textit{In-hospital event}. 
\end{itemize}

\
\setcounter{figure}{0}
\renewcommand{\figurename}{Figure}
\begin{figure*}[t]
    \includegraphics[width=1\linewidth]{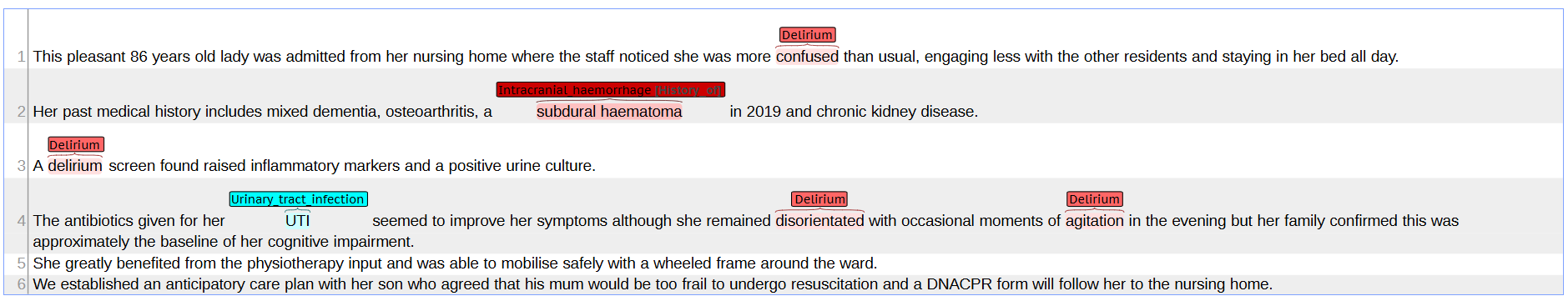}
    
    \caption{Example 3 on Brat}
    \label{example3}
\end{figure*}

\begin{figure*}[t]
    \includegraphics[width=1\linewidth]{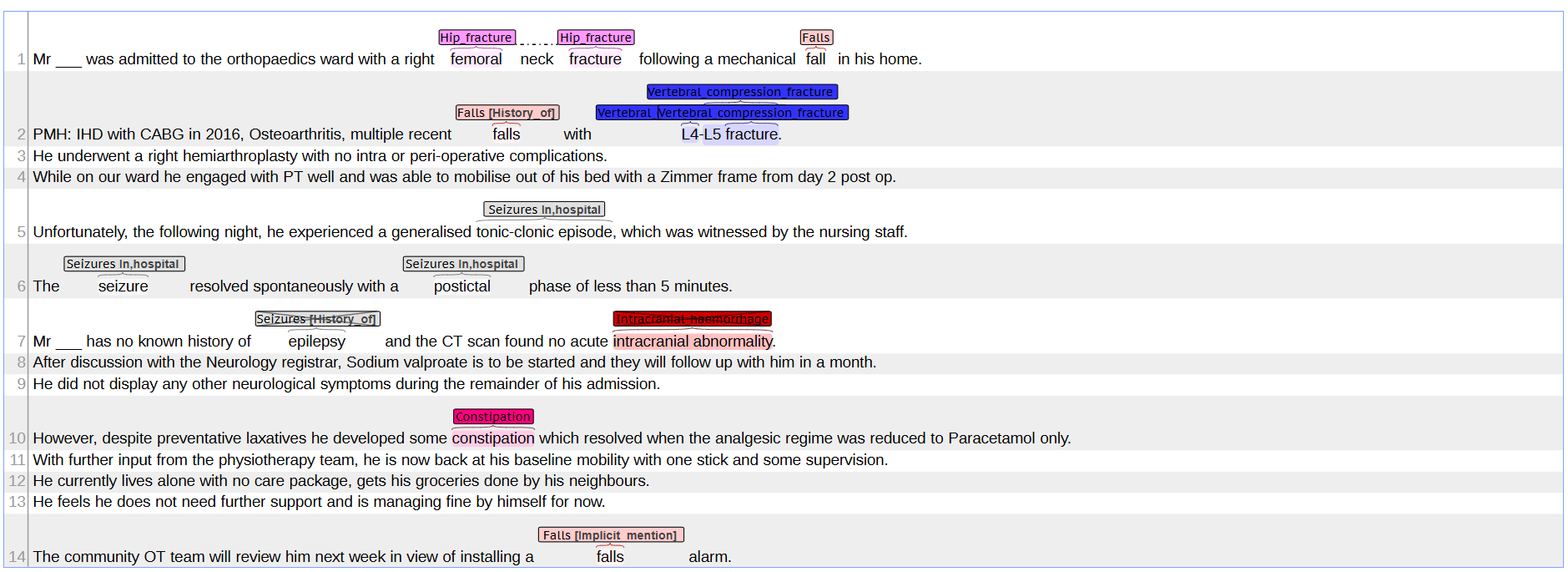}
    
    \caption{Example 4 on Brat}
    \label{example4}
\end{figure*}

\subsection{More examples}
\label{Appen4}
We are providing in the following section a set of examples for each AE in order to emphasise the different expressions that can be found in free text and how they should be annotated. We are also providing two more examples on Brat in Figures \ref{example3} and \ref{example4}.

\paragraph{Falls:} Different expressions can be used for falls including:
\begin{itemize}
    
    \item "\hl{Fell} down steps at front door last week."
    \item "Dizzy when turns quickly, \hl{fell} but only onto bed."
    \item "\hl{Found on floor} by carer"
    \item "His wife mentioned that she is very worried he could \hl{fall} when she is not watching him, the community occupational therapy team has planned to visit his house for a \hl{falls} assessment"
\end{itemize}
This sentence includes "Fall" and should be annotated. It describes initially a hypothetical event and later an assessment of the risk factors of falls in the patient's house. Both of these cases should be annotated as \textit{Implicit mention} because the geriatric syndrome has been described in the text but the event itself has not happened. 
\begin{itemize}
    \item "He started using a wheeled frame to avoid \hl{falling}."
\end{itemize}
This is another example where the patient has not fallen. Just like a risk of falls, this example should be annotated with the attribute \textit{Implicit mention}

Please note that falls should only be annotated in a geriatric context. For example, "fell off a wall while rock climbing" or "fell in horse riding accident" do not apply as these are not geriatric falls. Similarly, not all references to the noun or verb "fall" should be annotated. For example "falling sodium" or "patient moved house last fall" should be ignored.

\paragraph{Delirium} To annotate \textit{Delirium}, the annotator will be looking for evidence of acute fluctuation or resolution of the impaired cognition. For example:
\begin{itemize}
    \item "On admission, she was \hl{confused} and \hl{agitated} which settled to some extent with pain relief and rehydration."
\end{itemize}
This should be annotated as \textit{Delirium} even though it was not explicitly stated because we can see that the cognitive impairment has been resolved (albeit partially). Note that a precipitant can be identified in some cases (e.g. clues of bacterial infection alongside the cognitive impairment and improved cognition under antibiotics) and support the annotator's confidence. However, it is not a requirement for 10\%  of delirium diagnoses do not have a clear precipitant \cite{maclullich2009delirium}.

If the text suggests a cause for the patient's altered cognition which is not delirium, it should be ignored by the annotator, for example: 
\begin{itemize}
    \item "Patient was forgetful, disoriented: this was believed to be the result of metastatic disease to the brain."
\end{itemize}

\paragraph{Pressure injury:} Only ulceration or damage to the skin as a result of prolonged pressure should be included in this category. Other injuries such as venous ulcers and arterial ulcers should not be highlighted. If the type of injury is not explicitly named, its location\footnote{\url{https://www.cancerresearchuk.org/about-cancer/coping/physically/skin-problems/pressure-sores/causes-and-prevention}} is a strong marker, as shown in these examples:
\begin{itemize}
    \item "\hl{Pressure sore} right buttock, down to muscle" 
    \item "Known diabetic foot, skin break the right metatarsal head. Stage 3 \hl{bedsore} on left shoulder"
    \item "\hl{Hot spot} on left heel, deep ulcer right ankle"
\end{itemize}
An ulcer on the ankle is an unlikely location for a pressure injury and is presumably the result of venous insufficiency. It should therefore be ignored in the annotation.

\paragraph{Hip fractures:} As defined in Appendix 2, this label includes all proximal femoral fractures except isolated greater trochanter fractures and periprosthetic fractures. Different expressions can be found in free text to refer to hip fractures, including:
\begin{itemize}
    \item "\hl{Fracture} of the left neck of \hl{femur}" - In this example we want to include the term "fracture" as well as "femur" so we use a discontinuous entity to highlight both these elements as \textit{Hip fracture}.
    \item "He \hl{broke} his right \hl{hip} last year" - This is a discontinuous entity and should include the \textit{History of} attribute
    \item "Right intracapsular neck of \hl{femur fracture}" - This a straightforward example which does not require the use of discontinuous entities.
    \item "Pelvis and R Hip X-ray request: Ms\_\_\_ had a fall last night and landed on her right side. She was unable to stand and on examination, her left leg was shortened and externally rotated. \hl{NOF\#}?" - This is a radiology request (NOF means Neck of femur and \# means fracture) therefore this selection should be annotated with the \textit{Suspected} attribute.
    \item "Isolated fracture of the greater trochanter of the left femur." - This element should not be annotated. As described in Appendix 2, our definition of hip fractures excludes isolated greater trochanter fractures.
    \item "R \hl{Hip} X-ray: Some arthritic changes with narrowing of the joint space and osteophytes. No visible \hl{fracture}.” - This should be annotated with the discontinued entity \textit{Hip fracture} and the \textit{Negation} attribute.
    \item "This gentleman had a fall in his house, unable to weight bear since. ?Fracture of the left hip. Radiologists report: Left dynamic hip screw in position. No fracture observed." - We exclude in our annotation of hip fractures any periprosthetic fracture. Hence, the mention of a dynamic hip screw means this section of text should not be annotated.
\end{itemize}

\paragraph{Wrist fractures:} They are defined as fractures of the distal end of the radius, the distal end of the ulna or both. Different expressions can be found in free text to reference wrist fractures, including:
\begin{itemize}
    \item "\hl{Colles fracture} on the right side" - A Colles fracture is a specific fracture of the distal radius with dorsal angulation and displacement. It needs to be included in the selection because it locates the fracture on the wrist.
    \item "The X-ray showed an \hl{ulnar} styloid \hl{fracture}" - This is a discontinuous entity
    \item "She fell on her wrist and the X-ray showed a fracture of the 5th metacarpal" - This should not be annotated as we only want to identify fractures of the distal ulna and radius. 
    \item "Wrist X-ray report: Identifiable \hl{radial fracture} with posterior displacement."
    \item "The \hl{wrist} X-ray performed in A\&E showed no \hl{fracture}" - This is a discontinuous entity with the \textit{Negation} attribute
    \item "Imaging shows a right impacted and comminuted distal \hl{radial fracture} with associated avulsion \hl{fracture} of the \hl{ulnar} styloid process." - \textit{Radial fracture} is one continuous entity. And \textit{fracture} and \textit{ulnar} are a separate discontinuous entity.
\end{itemize}

\paragraph{Proximal humeral fractures:} Different expressions can be found in free text to reference proximal humeral fractures, including:
\begin{itemize}
    \item "X-ray confirmed a left \hl{humeral} neck \hl{fracture}"- This is a discontinuous entity
    \item "Her son reports that she had a \hl{broken shoulder} after a fall 6 months ago" - This is a discontinuous entity and should be annotated with the attributes \textit{History of}
\end{itemize}
Note that since the shoulder contains multiple bones and the terms "broken shoulder" are not specific to the proximal humerus, if the rest of the text does not specify its location any further, then this expression should be annotated with \textit{Low confidence}.
\begin{itemize}
    \item "CXR report: Both fields are clear, evidence of healed right neck of \hl{humerus fracture}."
\end{itemize}
Here the X-ray was performed to evaluate the patient's chest and not their shoulder or upper arm. Nonetheless, "humerus fracture" should be annotated along with the attribute \textit{History of}. This example will be detailed and annotated further in the \textit{Lower respiratory tract infection} paragraph.

\paragraph{Vertebral compression fractures:} The majority of vertebral fractures are the result of osteoporosis and unless the text mentions a risk of bony metastases from cancer or a high-velocity trauma, all occurrences of a fracture of vertebral body should be annotated with this entity. Fractures of the vertebral arch should be ignored. Most osteoporotic vertebral fractures will occur in the mid-thoracic and thoracolumbar regions below the T4 level\cite{griffith2015identifying} therefore any fractured vertebra above T4 level should not be annotated unless identified in the text as a result of osteoporosis. We recommend whenever possible to include the position of the fractured vertebra (T/L + number) in the selection of text. Different expressions can be found in free text to reference vertebral compression fractures, including:
\begin{itemize}
    \item "PMH: T2DM, \hl{L2 fracture}" - This also requires the \textit{History of} attribute.
    \item "Osteoporotic \hl{fracture} of \hl{L4}" - In this example, the highlighted text is one discontinuous entity.
    \item "Crush \hl{fractures} of \hl{T12} to \hl{L3} - In this example, we recommend using discontinuous entities and overlapping the selection of text so that two entities are extracted from this sentence: "fracture + T12" and "fracture + L3"
    \item "His wife mentioned that he was diagnosed with two \hl{broken vertebrae} in his lower back a month ago" - This would also require the \textit{History of} attribute.
    \item "X-ray revealed \hl{fractures} of \hl{T9 to 11}" - In this example, we can clearly annotate "fractures + T9" but annotating "fractures + 11" would not make sense because "T" is crucial to the site of the fracture. Therefore, our second selection should include "T9 to 11" to the fragment and this examples should be annotated as one discontinuous entity.
    \item "Lumbar Spine Xray: Image compared to December 2012. Noted \hl{vertebral fractures} of \hl{L4} and \hl{L5} present on previous images. Evidence of new \hl{fracture} of \hl{L4} with further loss of height" - In this example again we recommend annotating "vertebral fracture" and the use of discontinuous and overlapping entities for "fractures + L4" and "fractures + L5" alongside the \textit{History of} attribute. As for the last sentence of the example, we recommend selecting the text as one discontinuous entity and with no attribute.
\end{itemize}
This entity should also be used with the negation attribute whenever a Thoracic or Lumbar X-ray shows no fracture. For example:
\begin{itemize}
    \item "Lumbar Spine XR report: Multi level degenerative changes throughout the lumbar \hl{spine}. No \hl{fracture} detected." - This is one discontinuous entity with the \textit{Negation} attribute.
    \item "She was admitted with \hl{back} pain following a fall. X-ray was \hl{normal} and she progressed well with PT." - This is one discontinuous entity with the \textit{Negation} attribute.
\end{itemize}

\paragraph{Intracranial haemorrhage:} Intracranial haemorrhages are diagnosed on imaging of the head, most frequently used is the CT scan. Most CT scans of the head performed after a fall in the elderly or during an episode of delirium will be looking for an intracranial bleed. We therefore encourage the annotation of any normal CT scan result with this entity and the \textit{Negation} attribute. Different expressions can be found in free text to reference intracranial haemorrhage, such as:
\begin{itemize}
    \item "History of \hl{haemorrhagic stroke} at 52" - This should be annotated with the \textit{History of} attribute.
    \item "T2-FLAIR shows hyperdensity in keeping with \hl{subarchnoid haemorrhage} of left temporal region"
    \item "\hl{Head} CT found no mass or \hl{bleed}" - Both of these elements should be annotated as one discontinuous entity and with the \textit{Negation} attribute.
    \item "CT head shows acute \hl{subdural haematoma}" - The term "acute" should not be included in the selection because it does not help us identify the AE but simply indicates to us that this is a new or recent diagnosis.
    \item "CT head shows chronic \hl{subdural haematoma}" - The term "chronic" should not be included in the selection however the attribute \textit{History of} should be added to the annotation.
    \item "CT Head shows acute on chronic \hl{subdural haematoma}" - This selection of text should be highlighted twice, both with the entity \textit{Intracranial haemorrhage}, one with no \textit{Diagnosis} attribute and the other with the attribute \textit{History of}.
    \item "CT \hl{Head}: is \hl{unremarkable}" - Both of these elements should be annotated as one discontinuous entity and with the \textit{Negation} attribute.
    \item "CT Head: No \hl{intra-axial} or \hl{extra-axial haemorrhage}. No infarct visualised. Conclusion: No \hl{intracranial abnormality}." - This description of a CT scan with no intracranial haemorrhage requires three separate entities to be annotated. The first one is a discontinuous entity "intra-axial + haemorrhage", the second one is "extra-axial haemorrhage" (continuous) and the last one is "intracranial abnormality". Each of these elements of text should be annotated as \textit{Intracranial haemorrhage} with the \textit{Negation} attribute.
    \item "CT scan of head. Noted history of \hl{subdural haematoma} from a traumatic head injury six months ago, still present and stable on imaging. No new \hl{intracranial haemorrhage}" - The first element should be annotated with the attribute \textit{History of} and the second selection of text being preceded by "no new" reminds us that the patient had an intracranial bleed in the past while confirming there is no haemorrhage at present. Therefore it should be annotated once with the \textit{History of} attribute and a second time (selecting the exact same words) with the \textit{Negation} attribute.
    \item "CT shows an extensive \hl{cerebellar haemorrhage}"
\end{itemize}

\paragraph{Upper gastrointestinal haemorrhage:} The context around the portion of text we want to annotate might be useful to differentiate upper GI bleed from lower GI bleed:
\begin{itemize}
    
    \item "While in A\&E, she had \hl{coffee-ground emesis}"
    \item "He underwent emergency gastroscopy after 5 episodes of \hl{haematemesis} which found \hl{ruptured oesophageal varices}." - These are two distinct entities to be annotated with the same label
    \item "\hl{EOGD} found evidence of gastritis and no active \hl{bleeding}" - Discontinuous entity with the \textit{Negation} attribute
    \item "During the admission she had two episodes of \hl{melaena}"
\end{itemize}

\paragraph{Lower gastrointestinal haemorrhage:} Different expressions can be used for lower GI haemorrhage, including:
\begin{itemize}
    \item "I reviewed him at the surgery today for \hl{PR bleed}"
    \item "85-year-old male with a history of diverticulosis presented to the acute medical unit for new breathlessness on exertion and \hl{melaena}. Low HB was supplemented with 2 units of RBC, OGD was unremarkable, likely bleed from diverticulosis." - Melaena is usually a clinical sign of upper gastrointestinal haemorrhage but here the context indicates this is more likely a lower GI haemorrhage.
    \item "Reported \hl{diarrhoea} with mucus and \hl{blood}" - Discontinuous entity
    \item "The patient may notice traces of \hl{blood} in their \hl{stools} in the weeks following the surgery" - This is to be annotated as one discontinuous entity and with the \textit{Implicit mention} attribute
\end{itemize}

Despite their potential for detecting lower gastrointestinal haemorrhage, we recommend against the annotation of any bowel cancer screening tool such as faecal occult blood tests.

Please also note that clinicians might simply mention "gastro-intestinal haemorrhage" in the patient's records with no further specification. In this case, the annotator should highlight the expression with both entities \textit{Upper gastrointestinal haemorrhage} and \textit{Lower gastrointestinal haemorrhage} as well as the \textit{Low confidence} attribute.

\paragraph{Urinary tract infection:} For this entity, the annotator needs to find a mention of both the infection and its location on the urinary tract. Symptoms (frequency, burning sensation...), lab results (positive dip, cultures...) or treatment (antibiotics) should not be annotated. This is reflected in the examples below:
\begin{itemize}
    \item "Reason for admission: \hl{Pyelonephritis}"
    \item "Admitted with delirium. Urinalysis was positive and he was treated for a \hl{UTI} with good response."
    \item "Past medical history: Alzheimer's disease, GORD, Multiple \hl{CAUTI}" - The attribute \textit{History of} should be selected (CAUTI means catheter-associated urinary tract infection)
    \item "During this admission, he became more confused and his inflammatory markers raised. Urinalysis was positive, \hl{urine} culture was sampled and he started Nitrofurantoin based on previous sensitivities. The \hl{infection} settled after two days and he was able to be discharged back home with no further delay." - This should be annotated as one discontinuous entity and with the \textit{In-hospital event} attribute.
\end{itemize}
 Please note that a positive urine culture or urinalysis is not sufficient to confirm a urinary tract infection. Instead, the annotators should look for terms such as "infection" or "sepsis" and "urine" or "urinary tract", using the discontinuous entity tool when necessary.

\paragraph{Lower respiratory tract infection} Healthcare professionals often specify whether a lower respiratory tract infection was acquired in the community or in the hospital as this will guide the antibiotics they prescribe. If this is not explicit, the context around the annotated text can help guide the decision of the annotator to add the \textit{In-hospital event} attribute. Similarly to the entity \textit{Urinary tract infection}, the annotator needs to find a mention of both the infection (sepsis, septic, infective...) and its location on the lower respiratory tract (lung, chest, bronchial, pulmonary...). Symptoms (cough, shortness of breath...) , lab results (sputum cultures) and treatment (antibiotics) should be ignored. See the examples below for more details:
\begin{itemize}
    \item "New cough and wheeze during admission. Treated as hospital-acquired \hl{pneumonia}" - This should be annotated with the attribute \textit{In-hospital event}
    \item "Treated for ventilator-associated \hl{pneumonia}" - This should be annotated with the attribute \textit{In-hospital event}
    \item "Admitted from a care home for breathlessness, fever and chills. CXR was consistent with aspiration \hl{pneumonia}." - This should be annotated without the attribute \textit{In-hospital event}.
    \item "She received treatment 3 months ago for \hl{CAP}" - This should be annotated with the \textit{History of} attribute. "CAP" refers to "community-acquired pneumonia" so this should be annotated without the attribute \textit{In-hospital event}.
    \item "She received antibiotics post operatively for a \hl{HAP}" - Here "HAP" means hospital-acquired pneumonia and should be annotated with the \textit{In-hospital event} attribute.
    \item "CXR report: New \hl{consolidation} in the left lower lobe"
    \item "Reason for admission: \hl{Infective exacerbation} of \hl{COPD}" - Annotated as one discontinuous entity
    \item "Three episodes of vomiting in A\&E. Now coughing. CXR requested for ?\hl{aspiration}. - In this context, "aspiration" refers to "aspiration pneumonia" which is a lower respiratory tract infection following the inhalation of food, liquid or vomit. The annotation of the word "aspiration" on its own is recommended in this situation, alongside the \textit{Suspected} attribute
    \item "CXR confirms there is an \hl{infective} process in her right \hl{lung}" - One discontinuous entity
\end{itemize}
Since tuberculosis can manifest in various organs and systems within the body, we recommend that if the annotator selects it, they also add a location in the chest/lungs. For example
\begin{itemize}
    \item "CT scan shows a cavity in the apex of the left \hl{lung}. This appears to be scarring from previous \hl{TB} as demonstrated by previous imaging" - This would be annotated using the discontinuous entity tool and the attribute \textit{History of} should be added
\end{itemize}
The term "consolidation" when described in the report of an imaging of the chest almost always refer to an infection. When it does not, the radiologist or clinician will usually specify the differential diagnosis (e.g. "consolidation consistent with progression of the lung tumour"). Therefore, based on the context of the document, the annotators are encouraged to annotate this terms as \textit{Lower respiratory tract infection} if it appear to match this diagnosis.
\begin{itemize}
    \item "CXR showed a new patchy \hl{consolidation} of the right lower lobe. Antibiotics started for \hl{pneumonia}" - These are two separate entities annotated with the same label.
\end{itemize}
Most chest X-rays requested during the admission of an older adult will be looking for signs of an infection. Any normal chest X-ray will almost always exclude a lower respiratory tract infection. We therefore encourage annotators to highlight normal chest X-ray results with this entity and the negation attribute, as shown below:
\begin{itemize}
    \item "\hl{CXR} is \hl{normal}" - To be annotated as one discontinuous entity with the \textit{negation} attribute.
    \item "CXR report: Both \hl{lung} fields are \hl{clear}, evidence of healed right neck of humerus fracture." - For the \textit{Lower respiratory tract infection} entity, we want to annotate the above sections of text as one discontinuous entity with the \textit{Negation} attribute. This example was also reviewed in the \textit{Proximal humeral fractures} paragraph.
\end{itemize}
Note that the acronym "HCAP", or "HealthCare Associated Pneumonia" is often used by clinicians in their discharge letters. These describe a chest infection developed in the context of healthcare interventions such as an admission to the hospital or a stay in rehabilitation facility or long term care facility (e.g nursing home or care home). Our \textit{In-hospital event} attribute however, should only be used to identify GS or AE developed in hospital (acute or sub-acute care). Therefore a pneumonia diagnosed in the patient's records as "HCAP" simply because the patient was admitted from a nursing home or a care home should not be annotated with this attribute. On the other hand, an "HCAP" diagnosed part-way through the admission (not present when the patient initially entered the hospital) should be annotated as a \textit{Lower respiratory tract infection} with the \textit{In-hospital event} attribute.

\paragraph{Constipation:} Different expressions can be found in free text to refer to constipation, such as:
\begin{itemize}
    \item "\hl{Bowels not opened} in 5 days"
    \item "\hl{No recorded BM} in 4 days"
    \item "She reported a \hl{slower transit than usual}"
    \item "Adverse drug reactions: Amoxicillin (rash), Co-codamol (vomiting), Metformin (\hl{constipation})" - This should be annotated with the attribute \textit{History of}.
    \item "Discharge medications: Morphine sulphate 2mg every 4 hours as required, Senna 7.5 mg up to twice a day as required for \hl{constipation}" - In this context, the prescription of a laxative "as required" implies the patient might not be experiencing constipation. We recommend annotating any mention of the entity \textit{Constipation} with the attribute \textit{Implicit mention} when it appears in the medication list and described as "as required". This avoids extracting constipation when a laxative is given preventatively or absentmindedly left in a repeated prescription. If the patient reported constipation during their admission/presentation, the rest of the document would likely contain a mention of it and it would then be annotated without the attribute \textit{Implicit mention}.
\end{itemize}

\paragraph{Seizures:} Different expressions can be found in free text in reference to seizures, including:
\begin{itemize}
    
    \item "\hl{Postictal phase} inferior to 5min"
    \item "First \hl{grand mal} last month"
    \item "Family reported she had a \hl{fit} in the waiting room"
\end{itemize}

\section{More statistics about the annotated dataset}
\label{fine_grained_stat_fig}
\begin{figure*}[h]
    \includegraphics[width=1\linewidth]{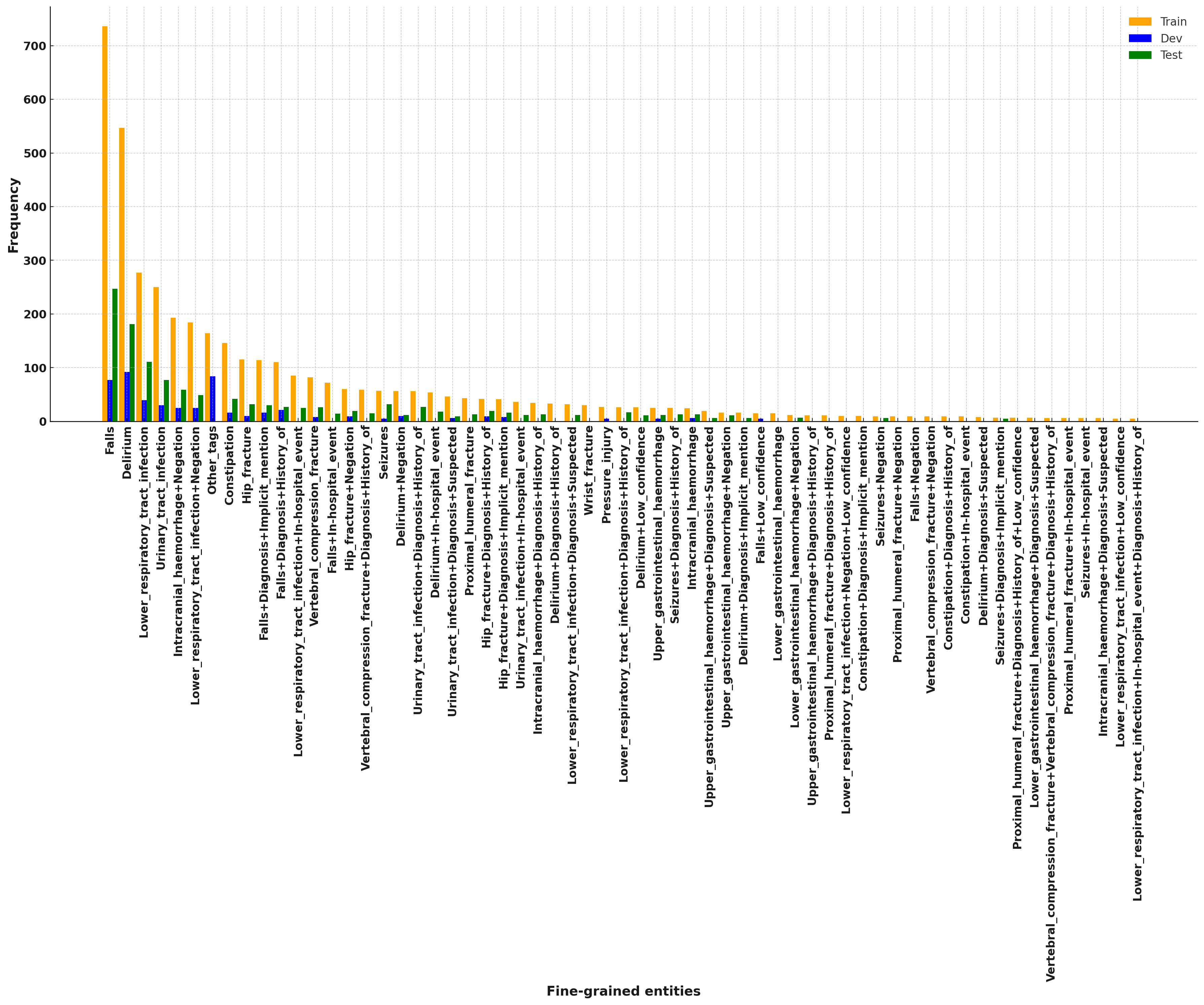}
    
    \caption{The distribution of fine-grained entities in the train, dev and test datasets}
    \label{fine_grained_stat}
\end{figure*}

Figure \ref{fine_grained_stat} illustrates the distribution of fine-grained entities in the train, dev and test corpora.
\section{Detailed results}
\label{resultsdetail}
\subsection{Best and worst examples for the entity level results (IAA)}
Table \ref{entity_best_worst} illustrates the five best and worst results (entity-level) related the agreement among the two annotator for both the pilot and the test corpora.
\setcounter{figure}{1}
\renewcommand{\figurename}{Table}
\begin{figure*}[h]
\begin{center}
     \begin{adjustbox}{width=1\textwidth}

\begin{threeparttable}
\caption{IAA between the two annotators (entity level) using bratiaa}
   \label{entity_best_worst}

\begin{tabular}{|p{1cm}|p{4cm}|p{9cm}|p{1.5cm}|p{10cm}|p{1.5cm}|}
\hline
\multicolumn{2}{|c}{}&\multicolumn{2}{|c|}{Best entities}&\multicolumn{2}{|c|}{Worst entities}\\
    \hline
   Corpus&Annotation type&Entity&f1-score&Entity&f1-score\\
   \hline
    \multirow{15}{*}{ Pilot}&\multirow{5}{*}{Fine-grained}&Delirium+Diagnosis+Suspected&1.000&Vertebral\_compression\_fracture+Diagnosis+History\_of&0.000\\
    \cline{3-6}
    &&Hip\_fracture+Diagnosis+History\_of&1.000&Constipation+Diagnosis+History\_of&0.000\\
    \cline{3-6}
    &&Seizures+Diagnosis+History\_of&1.000&Intracranial\_haemorrhage+Negation&0.000\\
    \cline{3-6}
    &&Upper\_gastrointestinal\_haemorrhage+Diagnosis+History\_of&1.000&Lower\_respiratory\_tract\_infection+Diagnosis+History\_of&0.000\\
    \cline{3-6}
    &&Urinary\_tract\_infection+Diagnosis+History\_of&1.000&Proximal\_humeral\_fracture+Negation&0.000\\
    \cline{2-6}
    &\multirow{5}{*}{Coarse-grained}&Pressure\_injury&1.000&Wrist\_fracture&0.000\\
    
    \cline{3-6}
    &&Falls&0.912&VIntracranial\_haemorrhage&0.000\\
    \cline{3-6}
    &&Upper\_gastrointestinal\_haemorrhage&0.875&Vertebral\_compression\_fracture&0.111\\
    \cline{3-6}
    &&Seizures&0.769&Proximal\_humeral\_fracture&0.222\\
    \cline{3-6}
    &&Urinary\_tract\_infection&0.760&Lower\_respiratory\_tract\_infection&0.596\\
    \cline{2-6}
    &\multirow{5}{*}{Coarse-grained+negation}&Pressure\_injury&1.000&Wrist\_fracture&0.000\\
    \cline{3-6}
  &&Upper\_gastrointestinal\_haemorrhage&0.933&Urinary\_tract\_infection+Negation&0.000\\
    \cline{3-6}
    &&Falls&0.912&Upper\_gastrointestinal\_haemorrhage+Negation&0.000\\
    \cline{3-6}
    &&Intracranial\_haemorrhage+Negation&0.750&Proximal\_humeral\_fracture+Negation&0.000\\
    \cline{3-6}
    &&Delirium+Negation&0.667&Lower\_respiratory\_tract\_infection+Negation&0.000\\
    
    \hline 
    \multirow{15}{*}{Test}&\multirow{5}{*}{Fine-grained}&Lower\_gastrointestinal\_haemorrhage+Diagnosis+History\_of&1.000&Seizures+Diagnosis+Referral&0.000\\
    \cline{3-6}
    &&Lower\_gastrointestinal\_haemorrhage+Diagnosis+Suspected&1.000&Wrist\_fracture+In-hospital\_event&0.000\\
    \cline{3-6}
    &&Intracranial\_haemorrhage+Diagnosis+History\_of&1.000&Vertebral\_compression\_fracture+Negation&0.000\\
    \cline{3-6}
    &&Hip\_fracture+In-hospital\_event&1.000&Urinary\_tract\_infection+Negation&0.000\\
    \cline{3-6}
    &&Falls+Negation+Diagnosis+History\_of&1.000&Upper\_gastrointestinal\_haemorrhage+Diagnosis+Suspected+In-hospital\_event&0.000\\
    \cline{2-6}
    &\multirow{5}{*}{Coarse-grained}&Wrist\_fracture&1.000&Vertebral\_compression\_fracture&0.488\\
    
    \cline{3-6}
    &&Urinary\_tract\_infection&0.927&Pressure\_injury&0.500\\
    \cline{3-6}
    &&Falls&0.909&Intracranial\_haemorrhage&0.523\\
    \cline{3-6}
    &&Seizures&0.885&Hip\_fracture&0.569\\
    \cline{3-6}
    &&Constipation&0.835&Upper\_gastrointestinal\_haemorrhage&0.697\\
    \cline{2-6}
    &\multirow{5}{*}{Coarse-grained+negation}&Proximal\_humeral\_fracture+Negation&1.000&Vertebral\_compression\_fracture+Negation&0.000\\
    \cline{3-6}
    &&Constipation+Negation&1.000&Urinary\_tract\_infection+Negation&0.000\\
    \cline{3-6}
    &&Lower\_gastrointestinal\_haemorrhage&1.000&Hip\_fracture+Negation&0.291\\
    \cline{3-6}
    &&Seizures+Negation&0.833&Intracranial\_haemorrhage+Negation&0.350\\
    \cline{3-6}
    &&Falls+Negation&0.800&Upper\_gastrointestinal\_haemorrhage+Negation&0.000\\
    
    \hline 
    
    \end{tabular}
 \begin{tablenotes}
 
 \item 
\end{tablenotes}
  \end{threeparttable}
\end{adjustbox}
\end{center}
\end{figure*}

\subsection{Best and worst examples for the document level results (IAA)}
Table \ref{doc_best_worst} illustrates the five best and worst results (document-level) related the agreement among the two annotator for both the pilot and the test corpora.
\begin{figure*}[h]
\begin{center}
     \begin{adjustbox}{width=1\textwidth}

\begin{threeparttable}
\caption{IAA between the two annotators (document level)}
   \label{doc_best_worst}

\begin{tabular}{|p{1cm}|p{4cm}|p{7cm}|p{1.5cm}|p{1.5cm}|p{8cm}|p{1.5cm}|p{1.5cm}|}
\hline
\multicolumn{2}{|c}{}&\multicolumn{3}{|c|}{Best entities}&\multicolumn{3}{|c|}{Worst entities}\\
    \hline
   Corpus&Annotation type&Entity&f1-score&Cohen’s kappa&Entity&f1-score&Cohen’s kappa\\
   \hline
    \multirow{15}{*}{ Pilot}&\multirow{5}{*}{Fine-grained}&Falls+Implicit\_mention&1.000&1.000&Falls+In-hospital\_event&0.000&0.000\\
    \cline{3-8}
    &&Delirium+Suspected&1.000&1.000&Upper\_gastrointestinal\_haemorrhage+Suspected&0.000&0.000\\
    \cline{3-8}
    &&Hip\_fracture+History\_of&1.000&1.000&Lower\_respiratory\_tract\_infection+In-hospital\_event&0.000&0.000\\
    \cline{3-8}
    &&Proximal\_humeral\_fracture+History\_of&1.000&1.000&Urinary\_tract\_infection+Suspected&0.333&0.296\\
    \cline{3-8}
    &&Falls+Diagnosis&0.800&0.779&Lower\_respiratory\_tract\_infection+Suspected&0.500&0.485\\
    \cline{2-8}
    &\multirow{5}{*}{Coarse-grained}&Pressure\_injury&1.000&1.000&Proximal\_humeral\_fracture&0.800&0.789\\
    
    \cline{3-8}
    &&Intracranial\_haemorrhage&1.000&1.000&Delirium&0.85&0.792\\
    \cline{3-8}
    &&Upper\_gastrointestinal\_haemorrhage&1.000&1.000&Hip\_fracture&0.860&0.846\\
    \cline{3-8}
    &&Constipation&1.000&1.000&Urinary\_tract\_infection&0.875&0.816\\
    \cline{3-8}
    &&Seizures&1.000&1.000&Lower\_respiratory\_tract\_infection&0.96&0.946\\
    \cline{2-8}
    &\multirow{5}{*}{Coarse-grained+negation}&Intracranial\_haemorrhage+Negation&1.000&1.000&Urinary\_tract\_infection+Negation:&0.000&0.000\\
    \cline{3-8}
    &&Delirium+Negation&1.000&1.000&Upper\_gastrointestinal\_haemorrhage+Negation&0.000&0.000\\
    \cline{3-8}
    &&Constipation&1.000&1.000&Proximal\_humeral\_fracture+Negation&0.000&0.000\\
    \cline{3-8}
    &&Seizures&1.000&1.000&Delirium&0.8333&0.781\\
    \cline{3-8}
    &&Proximal\_humeral\_fracture&1.000&1.00&Lower\_respiratory\_tract\_infection+B-Negation&0.857&0.847\\
    
    \hline 
    \multirow{15}{*}{Test}&\multirow{5}{*}{Fine-grained}&Lower\_gastrointestinal\_haemorrhage+Suspected&1.000&1.000&Intracranial\_haemorrhage+Suspected&0.000&0.000\\
    \cline{3-8}
    &&Proximal\_humeral\_fracture+Negation&1.000&1.000&Vertebral\_compression\_fracture+Negation&0.000&0.000\\
    \cline{3-8}
    &&Intracranial\_haemorrhage+History\_of&1.000&1.000&Hip\_fracture+In-hospital\_event&0.000&0.000\\
    \cline{3-8}
    &&Hip\_fracture+History\_of&0.963&0.929&Upper\_gastrointestinal\_haemorrhage+In-hospital\_event&0.000&0.000\\
    \cline{3-8}
    &&Intracranial\_haemorrhage+Diagnosis&0.923&0.857&Falls+Referral&0.000&0.000\\
    \cline{2-8}
    &\multirow{5}{*}{Coarse-grained}&Wrist\_fracture&1.000&1.000&Pressure\_injury&0.5&0.497\\
    
    \cline{3-8}
    &&Falls&0.989&0.983&Proximal\_humeral\_fracture&0.769&0.765\\
    \cline{3-8}
    &&Intracranial\_haemorrhage&0.966&0.961&Lower\_gastrointestinal\_haemorrhage&0.824&0.820\\
    \cline{3-8}
    &&Urinary\_tract\_infection&0.956&0.945&Constipation&0.836&0.821\\
    \cline{3-8}
    &&Lower\_respiratory\_tract\_infection&0.944&0.920&Hip\_fracture&0.841&0.817\\
    \cline{2-8}
    &\multirow{5}{*}{Coarse-grained+negation}&Constipation+Negation&1.000&1.000&Urinary\_tract\_infection+Negation&0.000&0.000\\
    \cline{3-8}
    &&Wrist\_fracture&1.000&1.000&Proximal\_humeral\_fracture+Negation&0.000&0.000\\
    \cline{3-8}
    &&Falls&0.989&0.984&Vertebral\_compression\_fracture+Negation&0.333&0.329\\
    \cline{3-8}
    &&Intracranial\_haemorrhage+Negation&0.978&0.976&Pressure\_injury&0.5&0.976\\
     \cline{3-8}
    &&Urinary\_tract\_infection&0.968&0.961&Upper\_gastrointestinal\_haemorrhage+Negation&0.615&0.610\\
   
    \hline 
    
    \end{tabular}
 \begin{tablenotes}
 
 \item 
\end{tablenotes}
  \end{threeparttable}
\end{adjustbox}
\end{center}
\end{figure*}

\subsection{Detailed results  (entity level)}
\begin{figure*}[h]
    \includegraphics[width=1\linewidth]{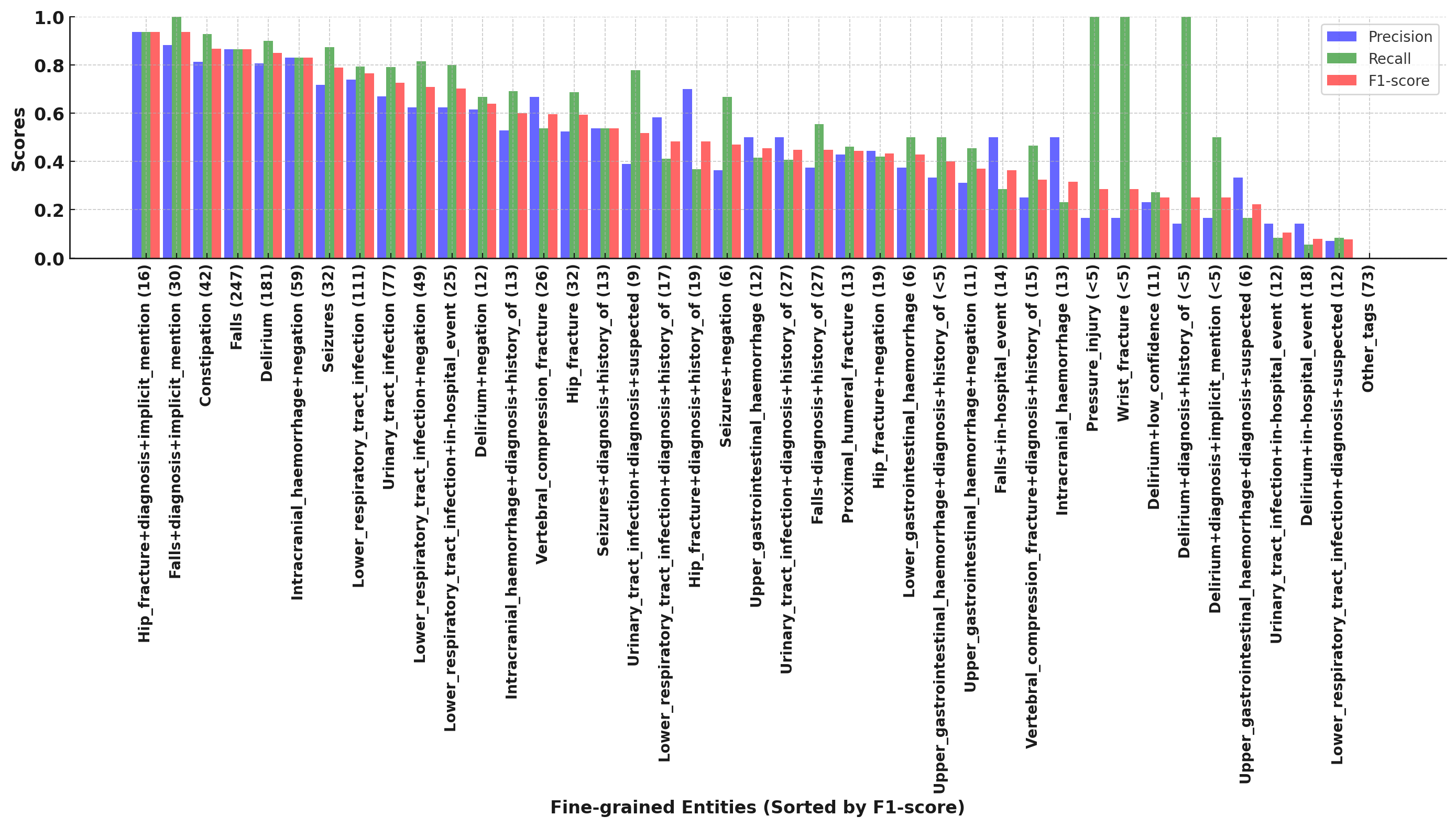}
    
    \caption{The results (precision, recall, F1-score) related to each fine\_grained entity within the test corpus (obtained with BioBERT)}
    \label{fine_grained_results}
\end{figure*}

\begin{figure}[h]
    \centering
    \includegraphics[width=1.2\linewidth]{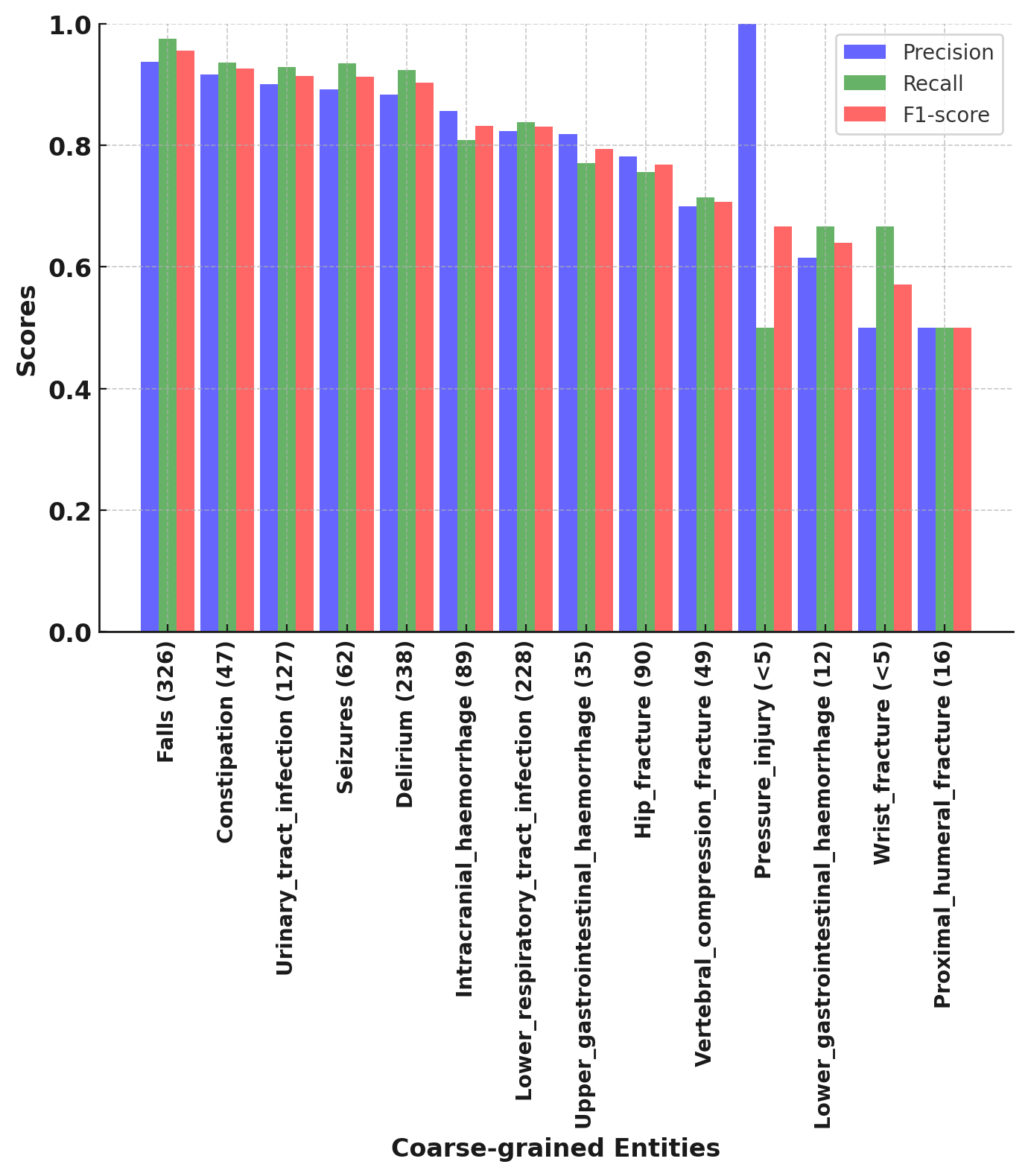}
    
    \caption{The results (precision, recall, F1-score) related to each coarse-grained entity within the test corpus (obtained with BERT\_cased)}
    \label{coarse_grained_results}
\end{figure}

\begin{figure}[h]
    \centering
    \includegraphics[width=1.2\linewidth]{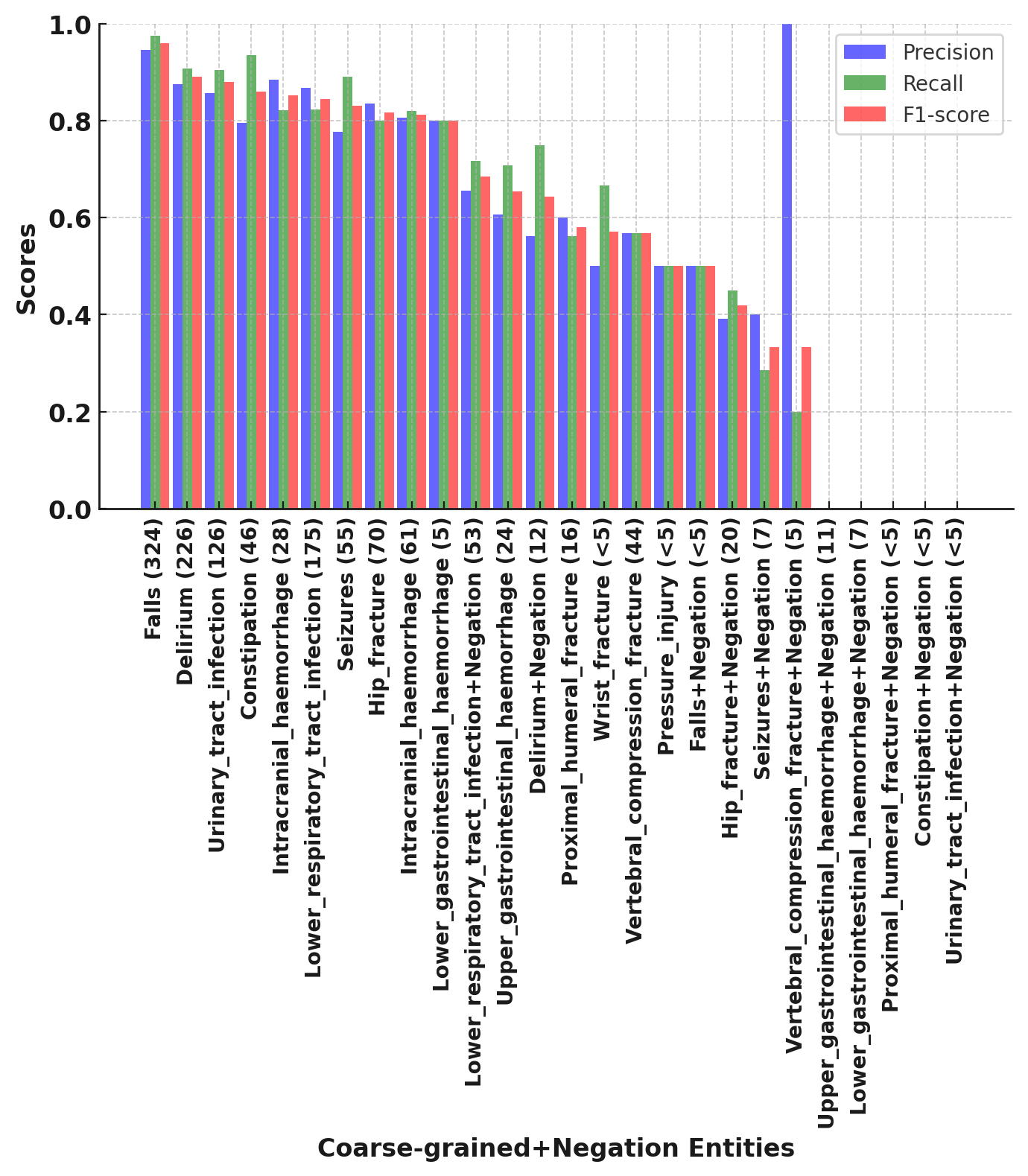}
    
    \caption{The results (precision, recall, F1-score) related to each coarse-grained+negation entity within the test corpus (obtained with BioBERT)}
    \label{coarse_grained_negation_results}
\end{figure}

Figures \ref{fine_grained_results}, \ref{coarse_grained_results} and \ref{coarse_grained_negation_results} illustrates the detailed precision, recall and f1-scores obtained for each entity in the dataset. We also highlight the frequency of each entity within the datset. For figure \ref{fine_grained_results}, we regroup all the entity having scores equal to 0 using the label "Other\_tags". In Other\_tags, we find rare entities such as \textbf{Lower\_respiratory\_tract\_infection+In-hospital\_event+Urinary\_tract\_infection+In-hospital\_even} or \textbf{Seizures+Diagnosis+History\_of+Negation}.

\renewcommand{\figurename}{Figure}

\section{More examples for the error analysis part}
\label{morebratexamples}

\renewcommand{\figurename}{Figure}

\begin{figure*}[h]
    \includegraphics[width=1\linewidth]{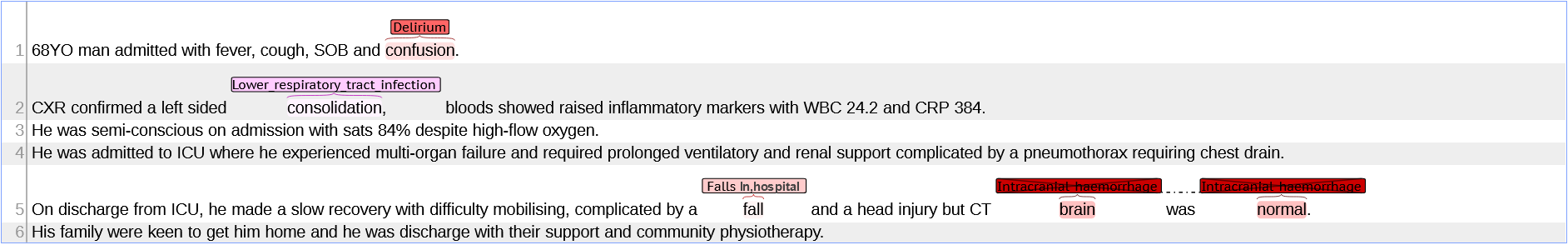}
    
    \caption{Example 2 on Brat}
    \label{example2_brat}
\end{figure*}

\begin{figure*}[h]
    \includegraphics[width=1\linewidth]{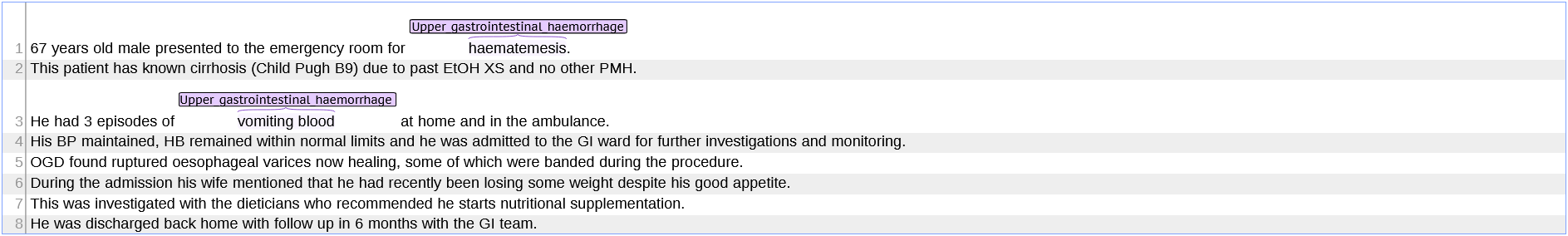}
    
    \caption{Example 3 on Brat}
    \label{example3_brat}
\end{figure*}
\setcounter{figure}{3}
\renewcommand{\figurename}{Table}
\begin{figure*}[t]
\begin{center}
     \begin{adjustbox}{width=1\textwidth}

\begin{threeparttable}
\caption{Some examples}
   \label{apply_model_example}

\begin{tabular}{|p{2cm}|p{4cm}|p{12.5cm}|p{7.5cm}|}
\hline
    Example&Annotation type&Entity level&Document level\\
   \hline
    \multirow{3}{*}{Example 2}&\multirow{4}{*}{Fine-grained}&\textbf{Span[10:11]}: confusion/Delirium(0.9996)&Delirium\\
    \cline{3-3}
    &&\textbf{Span[17:18]}: consolidation/Lower\_respiratory\_tract\_infection(1.0)&Lower\_respiratory\_tract\_infection\\
    \cline{3-3}
    &&\textbf{Span[48:49]}: fall/Falls(1.0)&Falls\\
    \cline{3-3}
    &&\textbf{Span[55:58]}: brain was normal/Intracranial\_haemorrhage+Negation(0.9992)&Intracranial\_haemorrhage+Negation\\
    \cline{2-4}
    
    &\multirow{4}{*}{Coarse-grained}&\textbf{Span[10:11]}: confusion/Delirium(1.0)&Delirium\\
    \cline{3-3}
    &&\textbf{Span[17:18]}: consolidation/Lower\_respiratory\_tract\_infection(1.0)&Lower\_respiratory\_tract\_infection\\
    \cline{3-3}
    &&\textbf{Span[48:49]}: fall/Falls(1.0)&Falls\\
    \cline{3-3}
    &&\textbf{Span[55:58]}: brain was normal/Intracranial\_haemorrhage(0.9998)&Intracranial\_haemorrhage\\
    \cline{2-4}
    (Figure \ref{example2_brat})&\multirow{4}{*}{Coarse-grained+negation}&\textbf{Span[10:11]}: confusion/Delirium(0.9698)&Delirium\\
    \cline{3-3}
    &&\textbf{Span[17:18]}: consolidation/Lower\_respiratory\_tract\_infection(1.0)&Lower\_respiratory\_tract\_infection\\
    \cline{3-3}
    &&\textbf{Span[48:49]}: fall/Falls(1.0)&Falls\\
    \cline{3-3}
    &&\textbf{Span[55:58]}: brain was normal/Intracranial\_haemorrhage+Negation(0.9979)&Intracranial\_haemorrhage+Negation\\
    \hline
    \multirow{3}{*}{Example 3}&\multirow{3}{*}{Fine-grained}&\textbf{Span[10:11]}: haematemesis/Upper\_gastrointestinal\_haemorrhage+ Diagnosis+History\_of(0.6954)&\\
    \cline{3-3}
    &&\textbf{Span[36:37]}: vomiting/Urinary\_tract\_infection+In-hospital\_event+ Lower\_respiratory\_tract\_infection+In-hospital\_event(0.0399)&Upper\_gastrointestinal\_haemorrhage\\
    \cline{3-3}
    &&\textbf{Span[37:38]}: blood/Lower\_respiratory\_tract\_infection+ Diagnosis+Suspected+Lower\_respiratory\_tract\_infection+Negation(0.0848)&\\
    
    \cline{2-4}
    &\multirow{1}{*}{coarse-grained}&\textbf{Span[10:11]}: haematemesis/Upper\_gastrointestinal\_haemorrhage(0.9994)&Upper\_gastrointestinal\_haemorrhage\\
    
    \cline{2-4}

    (Figure \ref{example3_brat})&\multirow{1}{*}{coarse-grained+Negation}&\textbf{Span[10:11]}: haematemesis/Upper\_gastrointestinal\_haemorrhage(0.9969)&Upper\_gastrointestinal\_haemorrhage\\
    \hline
    
    \end{tabular}
 \begin{tablenotes}
 
 \item 
\end{tablenotes}
  \end{threeparttable}
\end{adjustbox}
\end{center}
\end{figure*}

\end{document}